\documentclass{article}
\PassOptionsToPackage{numbers,sort&compress}{natbib}
\usepackage[main,final]{neurips_2026}

\usepackage{hyperref}
\usepackage{url}
\usepackage{times}

\title{Consolidating Reasoning with Test-Time Training}

\author{%
  Haotian Wu\thanks{Equal contribution. Correspondence to \texttt{zeming.chen@epfl.ch}} $\quad$ Zeming Chen\footnotemark[1] $\quad$ Hao Zhao $\quad$ Antoine Bosselut \\ \\
  EPFL
}

\usepackage[utf8]{inputenc}
\usepackage[T1]{fontenc}
\usepackage{nicefrac}
\usepackage{xcolor}
\usepackage{colortbl}
\usepackage{microtype}
\usepackage{hyperref}
\usepackage{url}
\usepackage{booktabs}
\usepackage{lineno}

\definecolor{darkblue}{rgb}{0, 0, 0.5}
\hypersetup{colorlinks=true, citecolor=darkblue, linkcolor=darkblue, urlcolor=darkblue}

\usepackage{amsmath}
\usepackage{amssymb}
\usepackage{amsfonts}
\usepackage{bm}

\usepackage{wrapfig}
\usepackage{graphicx}
\usepackage{caption}
\usepackage{subcaption}
\usepackage{multirow}

\usepackage{cancel}
\usepackage{placeins}
\usepackage{tikz}
\usetikzlibrary{arrows.meta,positioning}
\usepackage{pgfplots}
\usepackage{cleveref}

\DeclareMathOperator*{\argmax}{arg\,max}
\DeclareMathOperator*{\argmin}{arg\,min}

\newcommand{\answer}{y}
\newcommand{\canswer}{y^*}
\newcommand{\params}{{\bm{\theta}}}
\newcommand{\parallelttt}{\texttt{TTT-Consolidate}}
\newcommand{\openparallel}{\textit{OpenParallelThinking}}

\newcommand{\innerloss}{\mathcal{L}_{\text{inner}}}
\newcommand{\outerloss}{\mathcal{L}_{\text{outer}}}
\newcommand{\nllloss}{\mathcal{L}_{\text{NLL}}}

\newcommand{\metabaseparams}{\params_{\text{base}}}
\newcommand{\metaadapterparams}{\params_{\text{adapter}}}

\newcommand{\concat}{{\cdot}}

\NewDocumentEnvironment{alignb}{b}{%
  \begin{align*}
  \refstepcounter{equation} #1 \tag{\theequation}
  \end{align*}
}{\ignorespacesafterend}

\begin{document}

\maketitle

\begin{abstract}
Effective multi-agent collaboration requires finalizing solutions from reasoning that is distributed across the individual attempts of many agents. However, useful intermediate reasoning can remain unexploited when collaboration is limited to voting on conclusions reached among individual agents, or aggregating long, noisy collections of reasoning trajectories in context.
In this work, we propose \textbf{TTT-Consolidate}, a framework that uses multi-agent reasoning trajectories as adaptation data to improve a model’s subsequent reasoning. TTT-Consolidate jointly consolidates trajectories into parametric representations through test-time training, then generates a new solution by conditioning on the consolidated information. We meta-learn \textit{how to consolidate} using bi-level optimization at training time, where an inner loop encodes reasoning trajectories into a lightweight LoRA adapter, and an outer loop learns meta-parameters for the inner loop such that the consolidation is helpful for the model to synthesize a final solution.

As no large-scale corpus of diverse reasoning trajectories exists for training aggregation methods, we construct the \textit{OpenParallelThinking} corpus composed of 12,810 math problems paired with a mixture of trajectories from six LLMs. Evaluated on challenging math and STEM benchmarks, TTT-Consolidate outperforms the best aggregation baselines by up to 20 percentage points, even under matched compute and runtime settings, with the gap widening at larger model scales. TTT-Consolidate also recovers from unsuccessful individual trajectories more robustly, an ability that improves as the number of available (even incorrect) trajectories increases. Finally, we demonstrate that TTT-Consolidate benefits from both SFT and RL-based outer loop objectives. We will release TTT-Consolidate and \textit{OpenParallelThinking} for use in future work on reasoning aggregation.

\end{abstract}

\section{Introduction}

Multi-agent systems offer a promising way to extend the reasoning capabilities of large language models by allowing models to exchange, critique, and build on one another’s responses. Early work demonstrates benefits from multi-agent debate and layered synthesis of outputs from different models \citep{du2023debate,wang2024moa}. More recent work explores how learned teamwork strategies can combine partial reasoning into solutions absent from members' independent answers, and how communicating agents can build on intermediate discoveries while solving problems \citep{pappu2026sat,park2026communication}. Together, these directions motivate an important question: \emph{Can a model learn from other
agents' reasoning to construct a better solution?}

Effective collaboration requires aggregating distributed partial reasoning into a better solution by exploiting useful intermediate reasoning even when the attempts reach incorrect conclusions. However, common trajectory aggregation methods, such as majority voting, reward-guided selection, and best-of-$N$ sampling \citep{brown2024large,wang2023self, wu2025inference} do not effectively leverage all available reasoning signals in these trajectories. In effect, these methods treat each \textit{reasoning} trajectory independently, aggregate final \textit{answers} for a vote, and leave unexploited the potentially complementary reasoning encoded across the full ensemble of trajectories \citep{fu2025deepconf}. Although recent work proposes model-based in-context aggregation to better harness the reasoning across trajectories \citep{chen2023universal,qi2025learning,zhao2025majority}, these are attention-based aggregates. Studies of long-context language models show that access to information does not guarantee its effective use, and these methods still face typical long-context degradation issues \citep{hsieh2024ruler,kuratov2024babilong,dou2026clbench}.

We propose \parallelttt{}, a meta-learned framework that transforms trajectory aggregation from passive selection into active consolidation as test-time learning. At inference, \parallelttt{} consolidates diverse reasoning trajectories in a lightweight LoRA adapter via gradient updates and then synthesizes a final solution from the adapted model. We train the adapter via bi-level optimization; the inner loop minimizes a self-supervised next-token-prediction loss over a batch of trajectories, while the outer loop is trained to produce a synthesized answer, either through supervised finetuning on demonstrations distilled from a frontier teacher or through RLVR with GRPO \citep{shao2024deepseekmath, deepseekai2025deepseekr1incentivizingreasoningcapability}. The outer-loop objective backpropagates through the inner loop to shape how the LoRA aggregates the ensemble end-to-end, teaching the adapter to consolidate trajectories into representations that improve reasoning.

We train \parallelttt{} on \textit{OpenParallelThinking}, a novel dataset containing 13K math problems, each paired with groups of reasoning trajectories from 6 LLMs. Across math and STEM benchmarks, \parallelttt{} outperforms all rule- and model-based aggregation methods, with average gains of up to 20\% over the strongest baselines. These gains widen as model scale increases, and remain strong even under matched compute and runtime settings, which normally disfavor test-time training methods. Importantly, \parallelttt{} learns from the reasoning behavior in the trajectories, rather than shortcutting to voting based on the final answers in trajectories. In particular, on weak mixtures where no input trajectory is correct, \parallelttt{} can recover from these unsuccessful trajectories and arrive at an improved solution, outperforming the best baseline by up to 23\%, and its gains scale with the number of incorrect trajectories in consolidation. Finally, RL further improves the adapted policy's ability to reason over consolidated trajectories, adding an average 5-point gain over \parallelttt{} trained with SFT.

\section{Background and Preliminaries} \label{sec:background}

\begin{figure*}
    \centering
    \includegraphics[width=\textwidth]{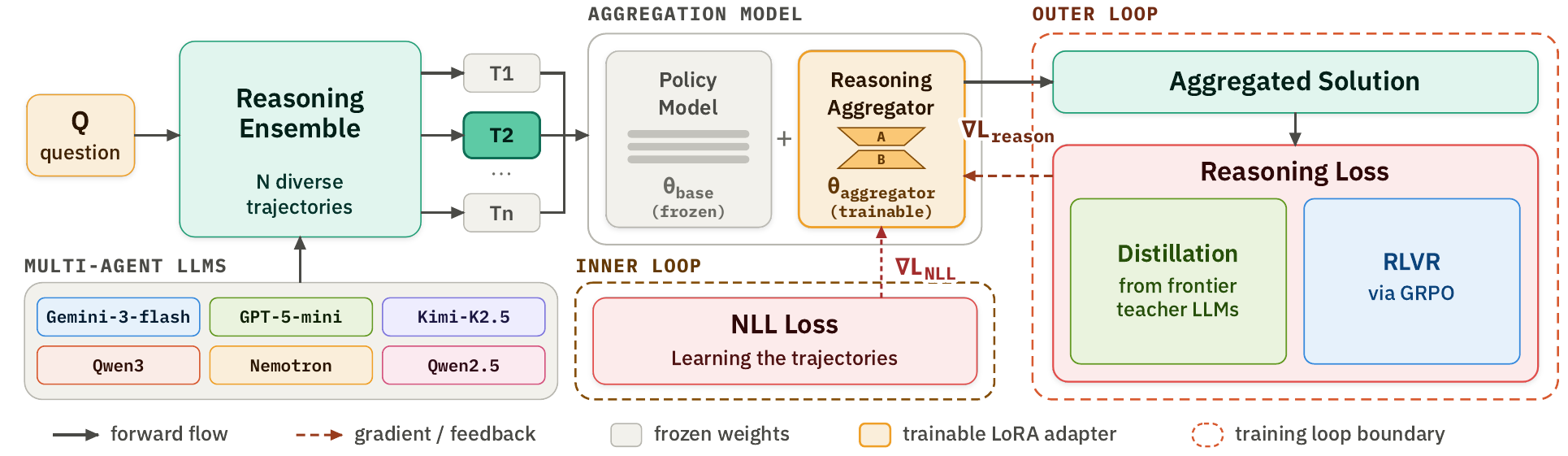}
    \caption{\textbf{Meta-learning \parallelttt{} for reasoning consolidation.} The training procedure involves a nested inner and outer loop. The inner loop optimizes the likelihood of a batch of reasoning trajectories with respect to the parameters of the LoRA reasoning aggregator. In the outer loop, the model uses the merged reasoning information in the aggregator to synthesize the final solution. In both loops, only the LoRA is updated while the base LLM is frozen.}
    \label{fig:overview}
\end{figure*}

\subsection{Test-Time Scaling and Reasoning Aggregation}

Scaling inference-time compute complements scaling parameters or pre-training data for improving LLM reasoning~\citep{wu2025inference}. Among test-time scaling strategies, \emph{parallel thinking}, i.e., sampling multiple reasoning trajectories and aggregating them into a single answer, has yielded remarkable empirical gains \citep{deepmind2025imo}. Formally, given a question $Q$ with verifiable ground-truth answer $\canswer$, parallel thinking involves sampling $N$ reasoning trajectories
$\mathcal{R} \,=\, \{R_1, \ldots, R_N\}$ from a causal language model $p_\params$ parameterized by $\params$, where each $R_n$ is a complete chain of thought ending in a candidate answer $a_n$. An \emph{aggregator} maps the question and trajectories to a final prediction. We write this as a generic operator: $\hat{\answer} \;=\; \mathcal{A}\bigl(Q,\, \mathcal{R};\, \params\bigr).$
The design space of $\mathcal{A}$ has evolved through three progressively more expressive paradigms.

\textbf{Rule-Based Aggregation.} $\;$
Standard aggregations apply hand-crafted rules to conclude a final answer from $R$. \emph{Majority Voting}~\citep{wang2023self} returns the modal answer, while \emph{reward-guided selection}~\citep{wu2025inference} ranks trajectories with an external scorer $r$ and keeps the top one:
\begin{equation}
\label{eq:rule-based}
\hat{\answer}_{\text{rule}} \;=\;
\begin{cases}
\displaystyle \mathrm{mode}\bigl(\{a_n\}_{n=1}^N\bigr) & \text{(majority vote)} \\[6pt]
\displaystyle a_{n^\star},\;\; n^\star \,=\, \argmax_{n\in[N]} r(Q,\, R_n) & \text{(reward selection)}
\end{cases}
\end{equation}
Rule-based aggregation discards intermediate reasoning, so partial correctness in trajectories with wrong final answers (e.g., a correct sub-derivation or a pruned but valid case) is never leveraged.

\textbf{Model-Based In-Context Aggregation.} $\;$
To exploit intermediate reasoning, recent work concatenates the trajectories into the prompt and produces an aggregated answer through a single forward pass with \emph{frozen} parameters~\citep{wan2025qwenlong,venkatraman2025recursive}:
$
\hat{\answer}_{\text{ctx}} \;\sim\; p_\params\!\bigl(\,\cdot \;\big|\; Q \concat R_1 \concat R_2 \concat \cdots \concat R_N\bigr).
$
Information from $\mathcal{R}$ enters the model only through self-attention over the context. As $N$ grows, the context length scales linearly, and LLMs are well documented to degrade on long, distractor-rich
inputs~\citep{hsieh2024ruler,kuratov2024babilong,xu2024stress,levy2024same,dou2026clbench},
leaving the rich complementary reasoning in $\mathcal{R}$ underexploited.

\textbf{RL-Trained Model Aggregation.} $\;$
A natural refinement keeps the in-context interface but \emph{trains} the aggregator end-to-end with reinforcement learning from verifiable rewards (RLVR), maximizing the expected reward of the aggregated output~\citep{zhao2025majority,venkatraman2025recursive}:
\begin{equation}
\label{eq:rl-agg}
\params^{\,\text{RL}} \;=\; \argmax_{\params}\;
\mathbb{E}_{(Q,\,\mathcal{R},\,\canswer)}
\Bigl[\,\mathbb{E}_{\hat{\answer}\,\sim\, p_\params(\,\cdot\,\mid\, Q,\,\mathcal{R})}\bigl[\,r(\hat{\answer},\, \canswer)\bigr]\Bigr].
\end{equation}
This turns aggregation into a learned skill rather than a fixed heuristic, but
still consumes $\mathcal{R}$ only through attention over a concatenated context, where the same long-context bottleneck persists.

\subsection{Test-Time Training}

A common thread in prior paradigms is that $\params$ is fixed at inference, and trajectories are processed only through context. In this work, we propose to perform aggregation with \emph{Test-time training} (TTT), consolidating the trajectories $\mathcal{R}$ directly in the model's parameters via $S$ gradient steps using a self-supervised language modeling objective $\nllloss$:
\begin{equation}
\label{eq:ttt}
\phi^{(s)} \;=\; \phi^{(s-1)} \,-\, \alpha\,\nabla_{\phi^{(s-1)}}\,
\nllloss\!\bigl(\phi^{(s-1)};\,\mathcal{R}\bigr),\qquad s = 1, \ldots, S,
\end{equation}
starting from $\phi^{(0)} = \params$ and yielding adapted parameters
$\phi(\params, \mathcal{R}) \triangleq \phi^{(S)}$. The aggregated final prediction
is then drawn from the \emph{updated} model:
$
\hat{\answer}_{\text{TTT}} \;\sim\; p_{\,\phi(\params,\,\mathcal{R})}\!\bigl(\,\cdot \,\mid\, Q\bigr).
$
This formulation is motivated by the observation that LLMs store and retrieve knowledge more reliably through their parameters than long-range attention~\citep{chen2025perk,tandon2025end,zweiger2025self}, sidestepping the long-context bottleneck that limits model-based aggregation.

\textbf{Meta-Learning for Test-Time Training.} $\;$
Training an LLM for TTT is naturally a meta-learning problem~\citep{finn2017maml, chen2023reckoning, vettoruzzo2024advances, chen2025perk}, where initial parameters $\params$ are meta-learned such that gradient-based adaptation of any $\mathcal{R}$ at test time yields strong downstream performance. With $\phi(\params, \mathcal{R})$ denoting the post-adaptation parameters, the meta-objective is the bilevel optimization
\begin{equation}
\label{eq:meta}
\params^{\star} \;=\; \argmin_{\params}\;
\mathbb{E}_{(Q,\,\mathcal{R},\,\canswer)}
\Bigl[\,\innerloss\!\bigl(\phi(\params, \mathcal{R});\; Q,\; \canswer\bigr)\Bigr],
\end{equation}
where $\innerloss$ governs how $\mathcal{R}$ is internalized and $\outerloss$ scores the adapted model on the downstream task. Following common practice~\citep{li2025beyond,chen2025perk,tandon2025end}, $\innerloss$ is the negative log-likelihood (NLL) over $\mathcal{R}$,
\begin{equation}
\label{eq:nll-inner}
\innerloss(\phi;\, \mathcal{R}) \;=\; -\,\sum_{k=1}^{N}\sum_{t=1}^{|R_n|} \log p_\phi\!\bigl(R_{n,t} \mid R_{n,<t}\bigr),
\end{equation}
which encodes the reasoning traces into the model's parameters. The outer-loop loss admits two natural instantiations:
\begin{equation}
\label{eq:outer-loss}
\outerloss(\phi;\, Q,\, \canswer) \;=\;
\begin{cases}
\displaystyle -\sum_{t=1}^{T_{\canswer}} \log p_\phi\!\bigl(\canswer_t \mid Q,\, \canswer_{<t}\bigr) & \text{(SFT)} \\[10pt]
\displaystyle -\,\mathbb{E}_{\,\answer\,\sim\, p_\phi(\,\cdot\,\mid\, Q)}\bigl[\,r(\answer,\, \canswer)\bigr] & \text{(RLVR)}
\end{cases}
\end{equation}
The SFT objective provides a dense, low-variance signal but requires annotated demonstrations, while the RLVR objective directly optimizes the verifiable performance at the cost of higher variance. \parallelttt{} builds on this formulation by instantiating the inner loop as a parameter-efficient (LoRA) adaptation that consolidates $\mathcal{R}$, and the outer loop as a learned synthesis of a final chain-of-thought (CoT) solution from the adapted model.

\section{\parallelttt: Reasoning Consolidation as Test-Time Learning}
\label{sec:method}

We instantiate the \textbf{Test-Time Training} (TTT) framework from Section~\ref{sec:background} for reasoning aggregation. Rather than treating $\mathcal{R}$ as a passive context, \parallelttt{} consolidates the trajectories into LoRA adapter weights $\phi$~\citep{hu2022lora} attached to a frozen backbone $\metabaseparams$ via gradient-based inner-loop adaptation, then answers $Q$ using the adapted model.
Figure~\ref{fig:overview} shows the procedure.

\textbf{Inner Loop Consolidation.} $\;\;$ For each training example
$(Q,\mathcal{R},\canswer)$, the inner loop initializes
$\phi^{(0)}=\metaadapterparams$ and adapts only $\phi$ while keeping
$\metabaseparams$ fixed:
\begin{equation}
\label{eq:inner}
\phi^{(s)}
=
\phi^{(s-1)}
-
\alpha
\nabla_{\phi}
\mathcal{L}_{\mathrm{inner}}
\!\left(\phi^{(s-1)};Q,\mathcal{R}\right),
\qquad
s=1,\ldots,S ,
\end{equation}
where the consolidation loss is the question-conditioned NLL of the sampled
reasoning trajectories,
\begin{equation}
\label{eq:consolidation-loss}
\mathcal{L}_{\mathrm{inner}}(\phi;Q,\mathcal{R})
=
-\sum_{k=1}^{K}
\sum_{t=1}^{|R_k|}
\log
p_{\metabaseparams\oplus\phi}
\!\left(R_{k,t}\mid Q,R_{k,<t}\right).
\end{equation}
Thus, the adapted $\phi^{(S)}$ serves as a parametric consolidation of $\mathcal{R}$. At inference time, the final answer is generated from $Q$ alone using $p_{\metabaseparams\oplus\phi^{(S)}}(\cdot\mid Q)$.

\textbf{Outer Loop Meta-Learning} $\;\;$ The outer loop optimizes $\metaadapterparams$ by differentiating the meta objective
through the inner-loop updates.
For SFT, we use the first branch of Eq.~\eqref{eq:outer-loss} with $p_\phi$ replaced by $p_{\metabaseparams\oplus\phi^{(S)}}$. The adapted model is trained using solution demonstrations from frontier teacher models, GPT-5.4-\texttt{xhigh} and Gemini 3.1 Pro Deep Think, which
provide high-quality CoT solutions as supervised targets~\citep{openai2026gpt54,google2026gemini31pro,google2026deepthink}.

For RLVR, we use the GRPO loss over the adapted policy (base model with adapted LoRA):
\begin{equation}
\label{eq:grpo-maml}
\begin{aligned}
\mathcal{L}_{\mathrm{GRPO\text{-}MAML}}
&=
-\frac{1}{G}
\sum_{i=1}^{G}
\frac{1}{|y_i|}
\sum_{t=1}^{|y_i|}
\min
\left(
r_{i,t}\hat{A}_i,\,
\operatorname{clip}(r_{i,t},1-\epsilon,1+\epsilon)\hat{A}_i
\right),
\\
r_{i,t}
&=
\frac{
p_{\metabaseparams\oplus\phi^{(S)}}
\!\left(y_{i,t}\mid Q,y_{i,<t}\right)
}{
p_{\metabaseparams\oplus\phi_{\mathrm{old}}^{(S)}}
\!\left(y_{i,t}\mid Q,y_{i,<t}\right)
}.
\end{aligned}
\end{equation}
Here $\{y_i\}_{i=1}^{G}$ are sampled responses from the adapted policy,
$\hat{A}_i$ are group-normalized advantages, and
$\phi_{\mathrm{old}}^{(S)}$ is the frozen reference copy of the adapted LoRA weights. To reduce memory, we apply truncated gradient unrolling (Appendix~\ref{app:subsection:implementation_details}).

\subsection{The \openparallel{} Dataset}

Existing reasoning datasets typically pair each problem with one solution, leaving little supervision for consolidating multiple reasoning traces. We address this gap with \openparallel{}, a dataset for training reasoning aggregators.

\begin{figure*}[!t]
    \centering
    \includegraphics[width=\linewidth]{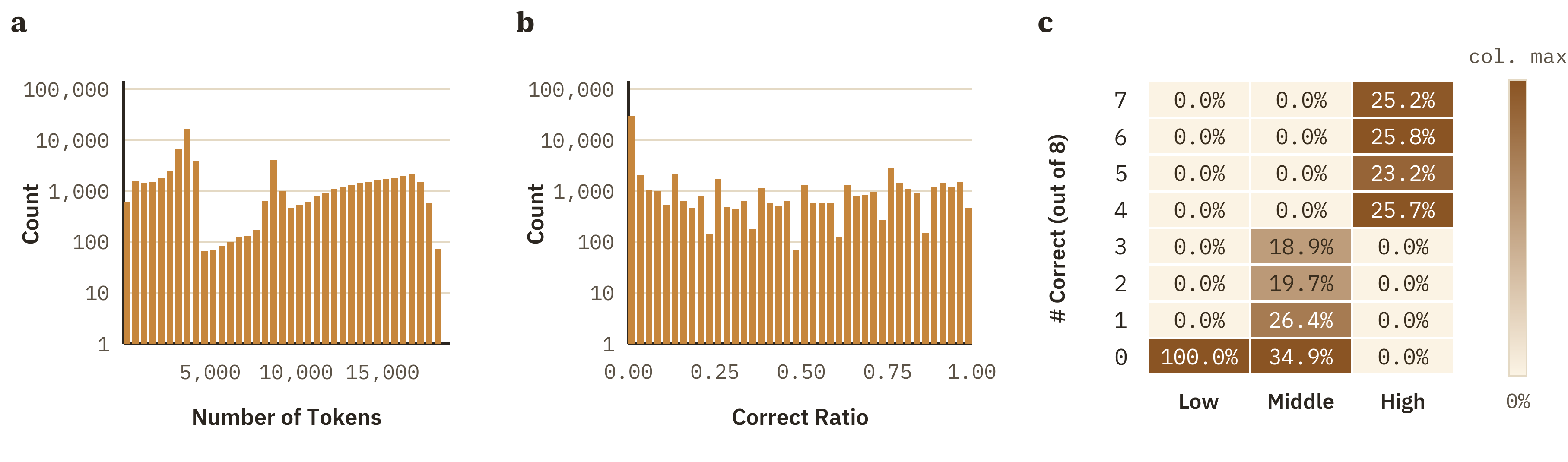}
    \caption{\textbf{Left:} Token length distribution of the \openparallel{}. \textbf{Center:} Correct ratio distribution \textit{OpenParallelThinking}, reflecting a cosine-like difficulty profile shaped by correctness-based filtering. \textbf{Right:} Correctness distribution of the evaluation trajectory groups across three tiers.}
    \label{fig:dataset_statistic}
\end{figure*}

\openparallel{} targets three design goals: source diversity, trajectory diversity, and balanced difficulty. We pool problems from four established math corpora and deduplicate them, then generate trajectories from four solvers spanning different scales, families, and training recipes, so the aggregator is not biased toward a single reasoning style. Trajectories are grouped per question under a token budget, and groups are filtered by correctness ratio to emphasize aggregation over predominantly incorrect trajectories, which yields a stronger learning signal than relying on a majority-correct shortcut (Figure~\ref{fig:dataset_statistic}). Appendix~\ref{app:subsection:openparallel_construction} gives the full construction pipeline, including source datasets, solver selection, deduplication, grouping, and filtering.
The final release contains 12,810 questions and 63,507 trajectory groups, drawn from 19,615 unique deduplicated problems. We sample 8,000 instances for RL training and reserve the rest for SFT. We will release the dataset and construction pipeline to support future work on reasoning aggregation.

\section{Experiments} \label{sec:experiments}

\subsection{Experimental Setup}

\paragraph{Benchmarks for Reasoning Aggregation.}
We evaluate on reasoning-heavy, meaningful benchmarks across two domains. For mathematical reasoning, we use AIME2025~\citep{aime2025}, AIME2026~\citep{aime2026}, Apex2025, ArxivMath, CMIMC2025, HMMT-feb-2026, and HMMT-nov-2025 from MathArena~\citep{balunovic_srimatharena_2025}, and IMO-Bench~\citep{luong-etal-2025-towards}. For generalization beyond math, we use two general STEM benchmarks: GPQA-diamond~\citep{rein2023gpqa} and TheoremQA~\citep{chen2023theoremqa}. For every problem, we generate eight reasoning trajectories (each ends with a final answer) distilled from LLMs spanning a range of capability levels. The generation and filtering are detailed in Appendix~\ref{app:subsection:reasoning_trajectory}. We group these trajectories into a quality bucket that isolates the effect of trajectory quality on aggregation performance: \textit{Minority} (0--3 correct). Essentially, having $\geq4$ correct trajectories already allows a simple rule-based aggregation like Majority Voting to get 100\% accuracy, which is unlikely to occur naturally. We additionally group trajectories into \textit{None}, where all trajectories are incorrect, to analyze whether \parallelttt{} can learn from their partial reasoning and aggregate them into a more reliable solution.

\paragraph{Baselines.}
We compare our method with four reasoning aggregation methods. For rule-based aggregation, we evaluate majority voting~\citep{wang2023self}. For model-based aggregation, we compare with three inference-time in-context aggregation variants: prompt aggregation, SFT aggregation, and RL aggregation ~\citep{zhao2025majority}.
Prompt aggregation uses the base model directly in a zero-shot manner, whereas SFT and RL aggregation use models trained with the corresponding objectives. We use Qwen3-1.7B and Qwen3-4B~\citep{yang2025qwen3} as the base models, and for all small datasets (less than 100 samples), and run them 8 times individually and report their average scores. For non-aggregation methods, we compare with the standard reasoning-heavy R1-style inference \citep{guo2025deepseek} with a 16K token extended thinking budget. See Appendix~\ref{app:subsection:implementation_details},~\ref{app:subsection:Prompts} for other implementation details.

\begin{table*}[t]
  \centering
  \setlength{\tabcolsep}{4pt}
  \renewcommand{\arraystretch}{1.05}
  \caption{\textbf{Main results (\%) on math and STEM reasoning benchmarks.}
    Each row group corresponds to a model scale and trajectory quality setting.
    \textit{None}: All 8 trajectories are incorrect.
    \textit{Minority}: A minority of trajectories are correct.
    \textbf{Bold} shows the best result per column within each group. \underline{Underline} reflects the best performance across all groups.
  }
  \label{tab:main}
  \resizebox{\textwidth}{!}{%
  \begin{tabular}{@{} c l *{8}{r} *{2}{r} @{}}
    \toprule
      & & \multicolumn{8}{c}{\textbf{Math Reasoning}}
      & \multicolumn{2}{c}{\textbf{STEM}} \\
    \cmidrule(lr){3-10}\cmidrule(lr){11-12}
    \textbf{Setting} & \textbf{Method}
      & \rotatebox{50}{\small \textbf{AIME25}} & \rotatebox{50}{\small \textbf{AIME26}} & \rotatebox{50}{\small \textbf{Apex25}}
      & \rotatebox{50}{\small \textbf{ArXivMath}} & \rotatebox{50}{\small \textbf{CMIMC25}} & \rotatebox{50}{\small \textbf{HMMT-2602}}
      & \rotatebox{50}{\small \textbf{HMMT-2511}} & \rotatebox{50}{\small \textbf{IMO-Bench}}
      & \rotatebox{50}{\small \textbf{GPQA}} & \rotatebox{50}{\small \textbf{TheoremQA}} \\[3pt]
    \midrule
    \multirow{2}{*}{\shortstack{Extended\\Thinking}}
      & Qwen3-1.7B
      & 40.0 & 40.8 & 0.0 & 13.9 & 25.3 & 21.9 & 25.8 & 22.8 & 37.4 & 57.5 \\
      & Qwen3-4B
      & \underline{71.7} & \underline{68.3} & 0.0 & 16.7 & 52.2 & 40.5 & 60.4 & 37.0 & 51.5 & \underline{68.6} \\
    \midrule
    \multirow{2}{*}{\shortstack{Rule-based\\Agg.}}
      & Majority Voting @8 \textbf{(None)}
      & 0.0 & 0.0 & 0.0 & 0.0 & 0.0 & 0.0 & 0.0 & 0.0 & 0.0 & 0.0 \\
      & Majority Voting @8 \textbf{(Minority)}
      & 26.7 & 23.3 & 25.0 & 18.1 & 30.0 & 30.3 & 16.7 & 21.5 & 16.2 & 19.5 \\
    \midrule
    \multirow{5}{*}{\shortstack{Qwen3-1.7B\\None}}
      & Prompt Aggregation\ (CoT)
      & 5.0 & 6.7 & 0.0 & 0.0 & 0.0 & 2.7 & 2.1 & 1.5 & 7.1 & 14.6 \\
      & Prompt Aggregation\ (Thinking)
      & {22.9} & {21.7} & 0.0 & {2.8} & {10.9}
      & {17.8} & {7.9} & {7.5} & {13.6} & {25.1} \\
      & SFT Aggregation\ (CoT)
      & 2.5 & 5.4 & 0.0 & 1.4 & 2.5 & 4.2 & 3.3 & 0.8 & 0.5 & 11.1 \\
      & RL Aggregation\ (Thinking)
      & 2.9 & 6.7 & 0.0 & {2.8} & 0.3 & 0.4 & 2.9 & 2.0 & 7.1 & 13.3 \\
    \rowcolor{orange!15}
      & \textbf{TTT-Consolidate}$^\dagger$
      & \textbf{36.6} & \textbf{26.7} & \textbf{16.7} & \textbf{23.6} & \textbf{20.0} & \textbf{18.2} & \textbf{23.3} & \textbf{14.5} & \textbf{22.7} & \textbf{33.5} \\
    \midrule
    \multirow{6}{*}{\shortstack{Qwen3-1.7B\\Minority}}
      & Prompt Aggregation\ (CoT)
      & 23.3 & 15.8 & 0.0 & 13.9 & 13.8 & 15.2 & 20.0 & 15.0 & 28.3 & 26.4 \\
      & Prompt Aggregation\ (Thinking)
      & {36.3} & \textbf{42.1} & {3.1} & 16.7 & {28.1}
      & {25.8} & 21.7 & {18.5} & {28.8} & {37.3} \\
      & SFT Aggregation\ (CoT)
      & 13.3 & 14.6 & 0.0 & 4.2 & 12.5 & 12.1 & 4.6 & 7.8 & 15.2 & 18.1 \\
      & RL Aggregation\ (Thinking)
      & 31.3 & 25.4 & 0.0 & {19.5} & 23.8
      & 22.4 & {23.8} & 17.3 & {28.8} & 27.3 \\
      & Recursive Refinement
      & 26.7 & 18.9 & 0.0 & 11.1 & 21.7 & 18.2 & 14.4 & 23.5 & 26.8 & -- \\
    \rowcolor{orange!15}
      & \textbf{TTT-Consolidate}$^\dagger$
      & \textbf{43.3} & {30.0} & \textbf{25.0} & \textbf{34.7} & \textbf{45.0} & \textbf{27.3} & \textbf{36.4} & \textbf{25.5} & \textbf{42.9} & \textbf{43.6} \\
    \midrule
    \multirow{5}{*}{\shortstack{Qwen3-4B\\None}}
      & Prompt Aggregation\ (CoT)
      & 11.7 & 4.6 & 0.0 & 4.2 & 4.4 & 9.1 & 6.7 & 3.0 & 10.1 & 21.3 \\
      & Prompt Aggregation\ (Thinking)
      & {32.9} & {36.7} & 0.0 & 5.6 & {24.5}
      & {24.6} & 14.2 & {12.0} &  {17.2} & {28.6} \\
      & SFT Aggregation\ (CoT)
      & 16.7 & 17.9 & 0.0 & 4.2 & 3.4 & 4.9 & 6.7 & 3.3 & 2.0 & 17.6 \\
      & RL Aggregation\ (Thinking)
      & 15.8 & 15.8 & {16.7} & {5.6} & 15.0
      & 9.1 & 20.0 & 6.7 & 4.3 & 32.6 \\
    \rowcolor{orange!15}
      & \textbf{TTT-Consolidate}$^\dagger$
      & \textbf{69.5} & \textbf{39.1} & \textbf{33.3} & \textbf{37.9} & \textbf{35.0} & \textbf{48.5} & \textbf{52.3} & \textbf{28.3} & \textbf{22.5} & \textbf{41.9} \\
    \midrule
    \multirow{5}{*}{\shortstack{Qwen3-4B\\Minority}}
      & Prompt Aggregation\ (CoT)
      & 28.3 & 26.7 & 0.0 & 16.7 & 31.9 & 25.0 & 26.7 & 18.0 & 29.3 & 33.4 \\
      & Prompt Aggregation\ (Thinking)
      & {61.7} & \textbf{61.7} & 2.1 & {23.6} & {44.1}
      & 26.1 & 35.8 & {23.8} & {30.8} & {39.8} \\
      & SFT Aggregation\ (CoT)
      & 30.0 & 23.3 & {8.3} & 6.9 & 15.0 & 18.2 & 20.0 & 7.0 & 3.5 & 22.0 \\
      & RL Aggregation\ (Thinking)
      & 56.7 & 53.3 & {8.3} & {23.6} & 40.0
      & {28.0} & {38.4} & 22.3 & 21.7 & 37.6 \\
    \rowcolor{orange!15}
      & \textbf{TTT-Consolidate}$^\dagger$
      & \textbf{69.6} & {47.6} & \underline{\textbf{41.6}} & \underline{\textbf{42.9}} & \underline{\textbf{78.6}} & \underline{\textbf{57.7}} & \underline{\textbf{72.7}} & \underline{\textbf{38.3}} & \underline{\textbf{64.6}} & \textbf{61.9} \\
    \bottomrule
  \end{tabular}%
  }
\end{table*}

\subsection{Main Results}

\textbf{\parallelttt{} consistently outperforms baseline aggregation methods.}
Across ten benchmarks and both trajectory-quality tiers, \parallelttt{} outperforms the baseline aggregation methods in nearly every case (Table~\ref{tab:main}).
At fixed model scale, the gain is stronger in the more difficult \textit{None} tier, where \parallelttt{}'s advantage grows precisely where input trajectories all contain incorrect conclusions and rule-based aggregation fails. At a fixed tier, scaling from 1.7B to 4B roughly doubles the gain, indicating that a stronger aggregator extracts more signal from the same set of trajectories. Compared to extended thinking with a 16K-token budget, \parallelttt{} trails by only 4.9 (1.7B) and 4.5 (4B) points when only aggregating incorrect trajectories (\textit{None}).
With \textit{Minority}, \parallelttt{} surpasses extended thinking by 7 (1.7B) and 10 (4B) points, with the gap widening at a larger scale. In contrast, every baseline aggregator trails extended thinking by a large margin, even in the \textit{Minority} setting.
Notably, SFT Aggregation is trained on the same questions, trajectory groups, and frontier-generated target solutions as \parallelttt{}, so this gap cannot be explained by frontier supervision in the outer loop alone: \parallelttt{}'s consolidation mechanism is what drives the improvement.

\begin{wrapfigure}[14]{r}{0.52\linewidth}
  \vspace{-0.2cm}
  \centering
  \includegraphics[width=\linewidth]{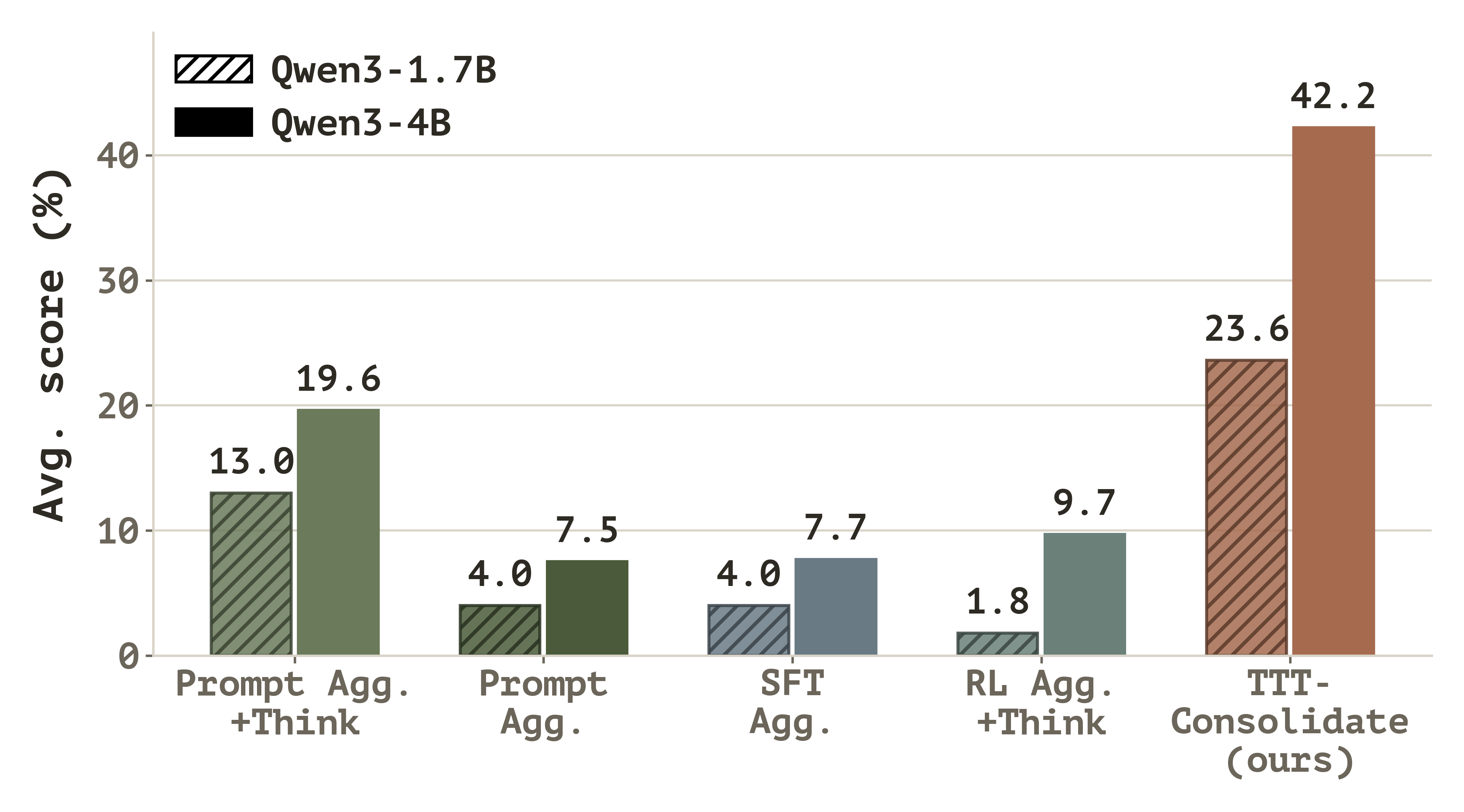}
  \caption{
    \textbf{Aggregating collectively incorrect trajectories (\emph{None}) with \parallelttt{}}. With both 1.7B and 4B models, \parallelttt{} improves the most compared to other methods.
  }
  \label{fig:none_quality_avg}
\end{wrapfigure}
\textbf{\parallelttt{} can aggregate unsuccessful trajectories into a more reliable solution.}
In the \textit{None} setting, every input trajectory arrives at an incorrect final solution, forcing majority voting to 0\% performance. Averaged across all ten benchmarks, \parallelttt{} achieves absolute gains of 22.6 and 10.6 points, approximately doubling the strongest baseline performance (Figure~\ref{fig:none_quality_avg}). The results suggest that test-time learning to consolidate can encode useful partial reasoning from collectively wrong attempts, producing answers stronger than individual attempts or the aggregation model solving the problem alone.\footnote{
  Qualitatively, we observe that \parallelttt{} can consolidate failed attempts into an improved solution. We use three frontier LLM judges (Fable~5, Opus~5, GPT-5.6-Sol; extra-high) to compare 100 \parallelttt{} solutions with their \textit{None} tier input trajectories. 98\% of solutions show clear synthesis of a better reasoning trajectory from the incorrect inputs. An audit of a separate 100 \textit{Minority}-tier solutions finds only 1 case of copying the input trajectory. Appendix~\ref{app:subsection:judge} shows the judge protocol and details.
}

\textbf{\parallelttt{} demonstrates out-of-domain transfer.}
\parallelttt{} is trained only on math reasoning from \textit{OpenParallelThinking} we constructed, but its advantage generalizes cleanly to scientific domains. Across GPQA and TheoremQA, \parallelttt{} averages 14 points above the strongest baseline across trajectory quality tiers and model scales, comparable in magnitude to its in-domain mathematical gains, suggesting consolidation based on test-time learning is a domain-general reasoning operation.

\section{Analysis} \label{sec:analysis}

\subsection{\parallelttt{}'s performance scales as the number of weak trajectories increases}
\label{sec:analysis:scaling}

\begin{wrapfigure}[13]{r}{0.58\linewidth}
  \vspace{-0.4cm}
  \centering
  \includegraphics[width=\linewidth]{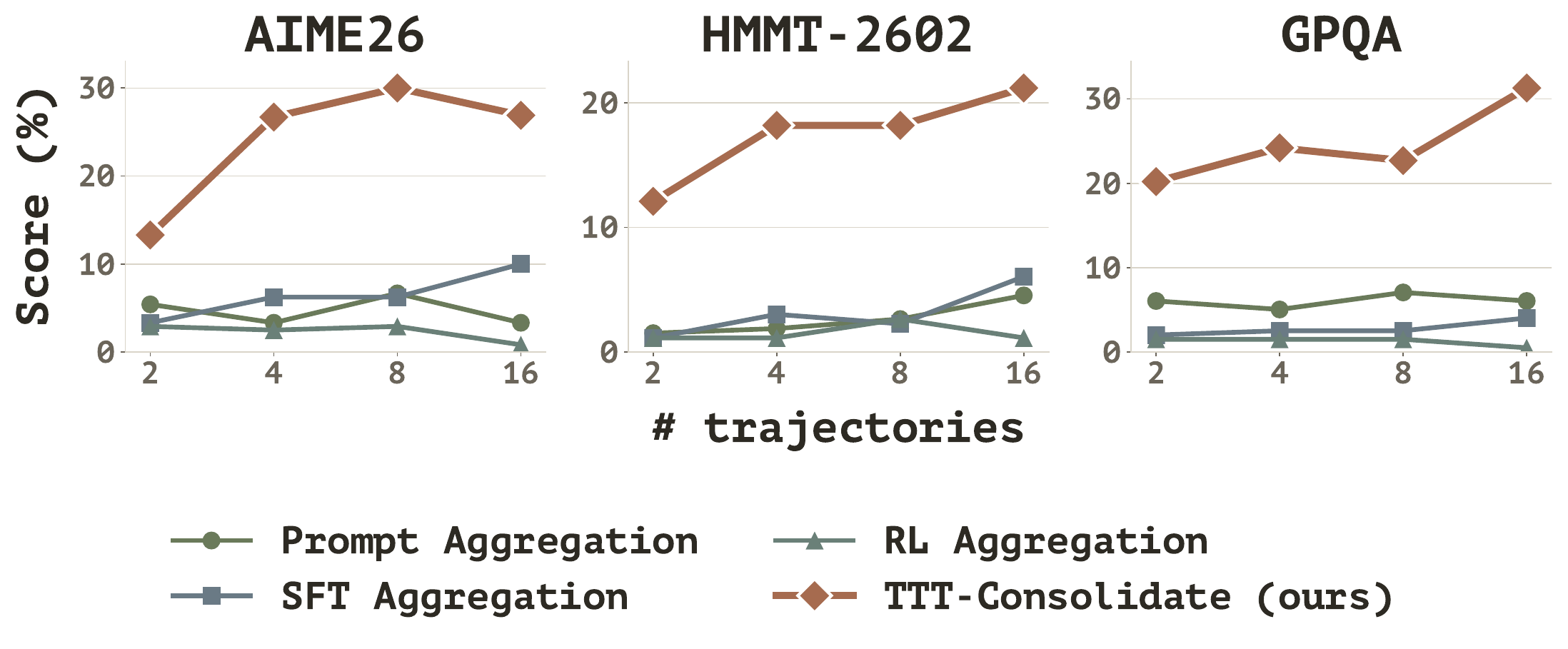}
  \caption{\textbf{Scaling the number of weak trajectories under aggregation.} As more \textit{None}-tier weak trajectories are aggregated, baseline aggregations improve slightly or degrade while \parallelttt{} continues to improve.}
  \label{fig:scaling_analysis}
\end{wrapfigure}

We analyze how aggregation methods scale with the number of weak trajectories under the \textit{None} tier. With all trajectories being incorrect, we rule out the shortcut of selecting an already-correct candidate, directly testing whether each method can extract useful knowledge from wrong attempts. We grow the number of trajectories as $n \in \{2, 4, 8, 16\}$, with Qwen3-1.7B as the backbone.

We show in Figure~\ref{fig:scaling_analysis} that as more incorrect trajectories are added, the baseline performance stagnates or even degrades. In contrast, \parallelttt{} generally benefits from more trajectories during consolidation, with, for instance, an 11.1\% increase on \texttt{GPQA}. We hypothesize that this stems from test-time learning deriving informative signal from noisy additional trajectories.

\subsection{\parallelttt{} outperforms compute-matched baselines}
\label{sec:analysis:compute}

\begin{wraptable}[12]{r}{0.55\linewidth}
  \vspace{-0.4cm}
  \centering
  \setlength{\tabcolsep}{4pt}
  \renewcommand{\arraystretch}{1.15}
  \caption{\textbf{Majority voting vs. \parallelttt{} on matched compute budget}. MV matches \parallelttt{}'s compute budget by spending on extra rollouts. MV does not improve with more rollouts and stays well below \parallelttt{}, even with more compute.}
  \label{tab:main_isocompute}
  \small
  \resizebox{\linewidth}{!}{
    \begin{tabular}{@{} l r r r @{}}
      \toprule
      \textbf{Method} & \textbf{FLOPs} & \textbf{Relative Compute} & \textbf{Minority Acc.} \\
      \midrule
      MV @8   & 7.019e14 & 0.67$\times$ & 22.7 \\
      MV @13  & 1.141e15 & 1.09$\times$ & 22.2 \\
      MV @18  & 1.579e15 & 1.50$\times$ & 22.6 \\
      \rowcolor{orange!15}
      \textbf{\parallelttt{} @8} & \textbf{1.050e15} & \textbf{1.00$\times$} & \textbf{35.4} \\
      \bottomrule
    \end{tabular}
  }
\end{wraptable}
\textbf{\parallelttt{} outperforms majority voting at matched compute.}
We verify if \parallelttt{}'s gain comes from the extra test-time compute by checking if a compute-matched baseline could achieve a similar performance. We allocate \parallelttt{}'s TTL budget to additional majority-voted rollouts (Table~\ref{tab:main_isocompute}).
Even with $1.50\times$ more FLOPs, majority voting @18 is still worse by 12.8 points than \parallelttt{}@8, and does not meaningfully improve as the compute grows (22.7 $\to$ 22.2 $\to$ 22.6). Appendix~\ref{app:compute} extends this comparison to reward-guided and recursive-refinement baselines on latency and memory.

\textbf{\parallelttt{} outperforms in-context aggregation even when using less compute.}
We further evaluate accuracy as a function of total computation, sweeping the number of incorrect trajectories $K$ (\textit{None}). We report average accuracy across the three benchmarks in Figure~\ref{fig:scaling_analysis} with the FLOPs for trajectory sampling and reasoning aggregation.
\begin{table*}[t]
  \centering
  \setlength{\tabcolsep}{4pt}
  \renewcommand{\arraystretch}{1.15}
  \caption{\textbf{Accuracy vs.\ compute} on the \textit{None} tier, averaged across the three benchmarks. $K$ denotes the number of input trajectories, all of which have incorrect final answers. We report accuracy (\%) and total FLOPs for trajectory sampling and aggregation. The in-context FLOPs apply to all in-context aggregation methods (prompt-only, SFT, RL). \parallelttt{} achieves higher accuracy even when the baselines use twice as many trajectories and more compute.}
  \label{tab:compute_vs_accuracy}
  \resizebox{0.98\textwidth}{!}{
    \begin{tabular}{@{} l r r r r r r r r @{}}
      \toprule
      \textbf{$K$}
        & \multicolumn{2}{c}{\textbf{MV}}
        & \multicolumn{4}{c}{\textbf{In-context Aggregation Methods}}
        & \multicolumn{2}{c}{\textbf{\parallelttt{}}} \\
      \cmidrule(lr){2-3}
      \cmidrule(lr){4-7}
      \cmidrule(lr){8-9}
        & \textbf{FLOPs} & \textbf{Acc.}
        & \textbf{FLOPs} & \textbf{Prompt Acc.}
        & \textbf{SFT Acc.} & \textbf{RL Acc.}
        & \textbf{FLOPs} & \textbf{Acc.} \\
      \midrule
      2  & 1.755e14 & 0.0 & 2.227e14 & 4.3 & 2.2 & 1.9 & 2.897e14 & \textbf{15.2} \\
      4  & 3.509e14 & 0.0 & 4.105e14 & 3.4 & 3.9 & 1.7 & 5.432e14 & \textbf{23.0} \\
      8  & 7.019e14 & 0.0 & 7.890e14 & 5.5 & 3.7 & 2.4 & 1.050e15 & \textbf{23.6} \\
      16 & 1.404e15 & 0.0 & 1.558e15 & 4.7 & 6.7 & 0.8 & 2.064e15 & \textbf{26.5} \\
      \bottomrule
    \end{tabular}
  }
\end{table*}
\parallelttt{} consistently outperforms every aggregation baseline even when they use $1.5\times$ more FLOPs and twice as many trajectories. \parallelttt{}'s performance increases with compute, from 15.2 to 26.5 when $K$ grows to 16. Majority voting always scores 0.0 since all candidates are wrong. Equal or greater FLOPs on additional sampling or in-context aggregation do not match \parallelttt{}'s performance. \parallelttt{}'s gains arise from consolidating useful intermediate reasoning through test-time learning, not simply from more test-time computation.

\subsection{\parallelttt{} requires joint consolidation in parameter space}
\label{sec:analysis:single_trace}
\begin{figure*}[t]
  \centering
  \definecolor{cbGray}{HTML}{9A9590}
  \definecolor{cbOrange}{HTML}{E69F00}
  \definecolor{cbSkyBlue}{HTML}{56B4E9}
  \definecolor{cbGreen}{HTML}{009E73}
  \definecolor{cbYellow}{HTML}{F0E442}
  \definecolor{cbTextPrimary}{HTML}{2D2A23}
  \definecolor{cbArrow}{HTML}{6B655A}
  \resizebox{\textwidth}{!}{
  \begin{tikzpicture}[
      font=\tiny,
      every path/.style={draw=cbArrow},
      box/.style={rounded corners, minimum height=4mm, minimum width=7mm, align=center, inner sep=1.5pt, font=\ttfamily\bfseries\tiny, text=cbTextPrimary},
      traj/.style={rounded corners, minimum height=4mm, minimum width=4mm, align=center, inner sep=1pt, font=\ttfamily\bfseries\tiny, text=cbTextPrimary, fill=cbGray!25},
      lora/.style={box, fill=cbOrange!35},
      outbox/.style={rounded corners, minimum height=4mm, minimum width=4mm, align=center, inner sep=1pt, font=\ttfamily\bfseries\tiny, text=cbTextPrimary, fill=cbSkyBlue!30},
      ttlicon/.style={circle, fill=cbOrange!90, inner sep=0pt, minimum size=3mm, font=\ttfamily\fontsize{3.6}{3.6}\selectfont, text=black},
      -, shorten >=1pt,
    ]
    \begin{scope}
      \node[font=\bfseries\ttfamily\tiny, text=cbTextPrimary] (title1) at (1.55,1.85) {Single-Trace$\times$8};
      \node[traj] (r1) at (0,1.35)  {$R_1$};
      \node[traj] (r2) at (0,0.9)  {$R_2$};
      \node[font=\tiny] (rd) at (0,0.5) {$\cdots$};
      \node[traj] (r8) at (0,0.1)  {$R_8$};

      \node[ttlicon] (g1) at (0.7,1.35) {$\boldsymbol{\nabla}$};
      \node[ttlicon] (g2) at (0.7,0.9) {$\boldsymbol{\nabla}$};
      \node[ttlicon] (g8) at (0.7,0.1) {$\boldsymbol{\nabla}$};

      \node[lora] (l1) at (1.4,1.35) {LoRA$_1$};
      \node[lora] (l2) at (1.4,0.9) {LoRA$_2$};
      \node[lora] (l8) at (1.4,0.1) {LoRA$_8$};

      \node[outbox]  (y1) at (2.1,1.35) {$y_1$};
      \node[outbox]  (y2) at (2.1,0.9) {$y_2$};
      \node[outbox]  (y8) at (2.1,0.1) {$y_8$};

      \node[box, fill=cbGreen!25] (vote) at (3.1,0.725) {Majority\\Vote};

      \draw (r1)--(g1); \draw (g1)--(l1); \draw (l1)--(y1); \draw (y1)--(vote);
      \draw (r2)--(g2); \draw (g2)--(l2); \draw (l2)--(y2); \draw (y2)--(vote);
      \draw (r8)--(g8); \draw (g8)--(l8); \draw (l8)--(y8); \draw (y8)--(vote);
      \node[align=center, text width=3.6cm] at (1.55,-0.55) {No interaction between trajectories during adaptation.};
    \end{scope}

    \begin{scope}[xshift=4.4cm]
      \node[font=\bfseries\ttfamily\tiny, text=cbTextPrimary] (title2) at (1.6,1.85) {Full \parallelttt{}};
      \node[traj] (s1) at (0,1.35) {$R_1$};
      \node[traj] (s2) at (0,0.9) {$R_2$};
      \node[font=\tiny] (sd) at (0,0.5) {$\cdots$};
      \node[traj] (s8) at (0,0.1) {$R_8$};

      \node[ttlicon] (joint)  at (0.8,0.725) {$\boldsymbol{\nabla}$};
      \node[lora] (shared) at (1.9,0.725) {Consolidate\\LoRA};
      \node[outbox]  (sol)    at (3.3,0.725) {$y$};

      \draw (s1)--(joint); \draw (s2)--(joint); \draw (s8)--(joint);
      \draw (joint)--(shared); \draw (shared)--(sol);
      \node[align=center, text width=4.0cm] at (1.7,-0.55) {Interaction occurs before decoding, in parameter space.};
    \end{scope}

    \begin{scope}[xshift=8.2cm]
      \node[font=\bfseries\ttfamily\tiny, text=cbTextPrimary] (title3) at (1.5,1.85) {Result};
      \node[align=center] (resbar) at (1.5,0.7) {\includegraphics[height=1.5cm]{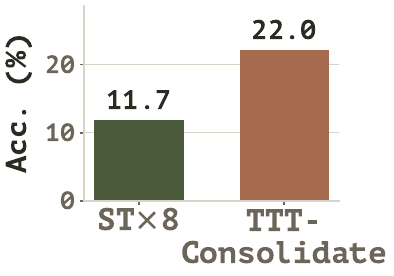}};
      \node[align=center, text width=3.6cm] at (1.5,-0.55) {Joint consolidation gains $+10.3$ points at matched compute.};
    \end{scope}
  \end{tikzpicture}}
  \caption{\textbf{Does trajectory information interact in the consolidation step?} Under Single-Trace$\times$8 (left), each trajectory is adapted and decoded independently, then reconciled by a vote. Under full \parallelttt{} (middle), all 8 trajectories jointly update one LoRA before any answer is generated. Single-Trace$\times$8 underperforms full \parallelttt{} on the \textit{None} tier (right).}
  \label{fig:single_trace_diagram}
\end{figure*}

We validate that \parallelttt{}'s gains require cross-trajectory interaction, not repeated exposure to the same question or task-distribution activation. We compare \parallelttt{} to using 8 separately adapted LoRAs, one per input trajectory, with a majority vote over their 8 generated solutions (\textbf{Single-Trace$\times$8}; Figure~\ref{fig:single_trace_diagram}). Both see the same 8 trajectories and spend matched adaptation compute. The only difference is whether cross-trajectory interaction happens before producing a solution.
Single-Trace$\times$8 is worse by 10.3 points at matched compute, showing that \parallelttt{}'s gain requires consolidating trajectories \emph{jointly} within a shared parameter space (adapted LoRA) rather than encoding them independently and reconciling only by voting over outputs. Appendix~\ref{app:ablation:controls} extends this comparison with two more controls, passive question repetition and task-distribution activation without target-specific trajectories, both of which perform worse than Single-Trace$\times$8, ruling them out as alternative explanations.

\subsection{\parallelttt{} consolidates reasoning rather than voting on the answers}
\label{sec:analysis:shortcuts}

\begin{wrapfigure}[22]{r}{0.5\linewidth}
  \vspace{-0.4cm}
  \centering
  \includegraphics[width=\linewidth]{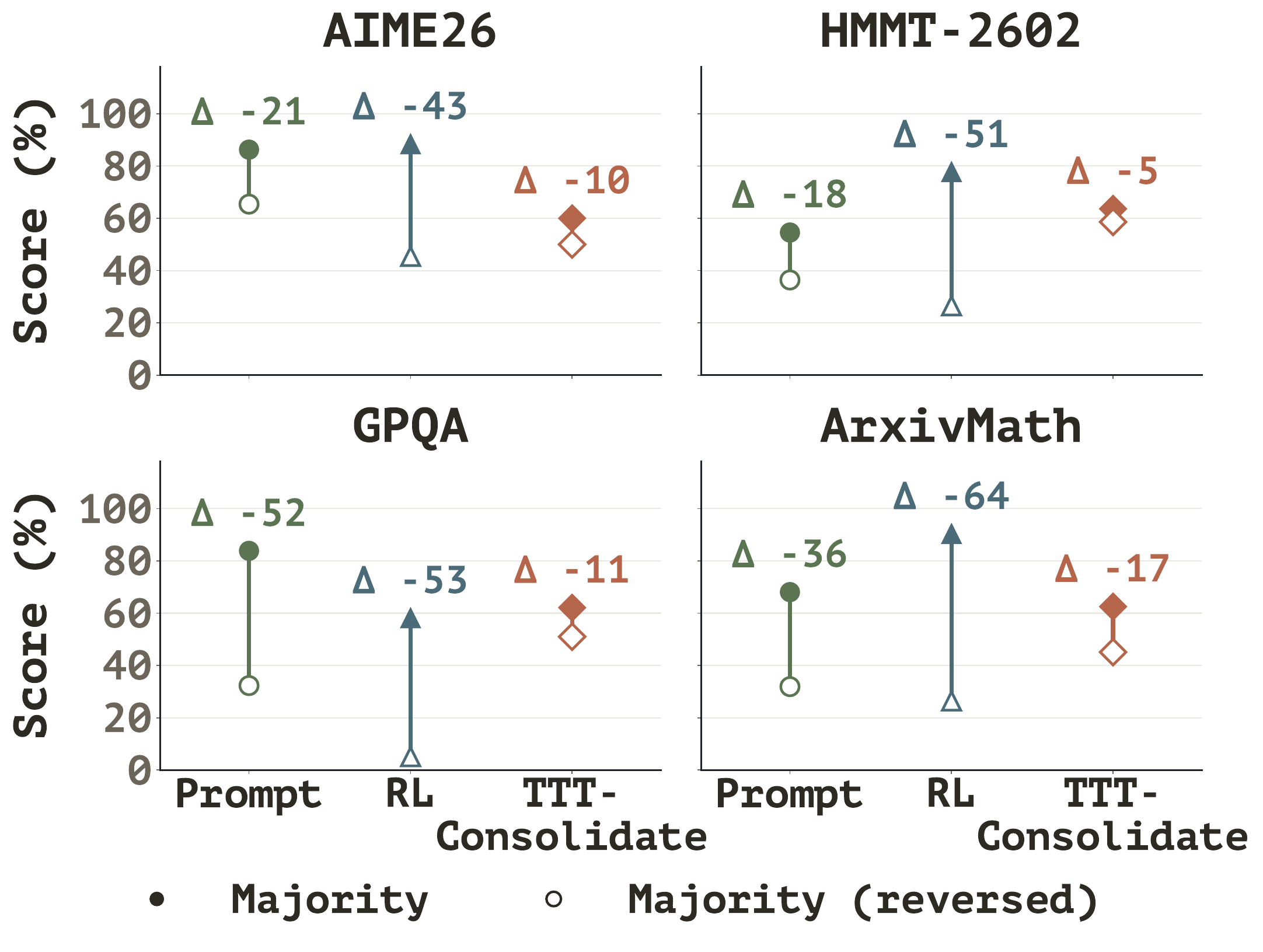}
  \caption{\textbf{Robustness to majority manipulation.} Dumbbells connect each method's score under \emph{Majority} (filled) and \emph{Majority-reversed} (open, correct/incorrect answers swapped); bar length is the resulting drop. Baselines collapse while our method stays short in every panel, indicating consolidation rather than answer-tallying.}
  \label{fig:majority-reversal-dumbbell}
\end{wrapfigure}

An aggregation method that scores well when most trajectories carry the correct answer could either be consolidating reasoning or simply voting on final answers. To distinguish these, we use the \textit{Majority} quality bucket: 4 to 7 trajectories carry correct answers. We swap the correct/incorrect answers so the majority answer becomes wrong, while reasoning stays untouched (\textit{Majority-reversed}). A consolidator should be largely invariant, but a ``voter'' should collapse. We show the results in Figure \ref{fig:majority-reversal-dumbbell}.

Under the \textit{Majority} tier, Prompt and RL Aggregation reach 80–90\% on several benchmarks. However, after reversal, RL drops by up to 64 points and Prompt by up to 52 points, often falling below 30\%. The large drop suggests their high scores came from shortcut voting on final answers rather than understanding the trajectories. In contrast, \parallelttt{}'s drop stays within the $[-17, -5]$ range across all benchmarks, indicating it utilizes information from trajectories rather than counting answers.

\newpage
\subsection{RL incentivizes \parallelttt{}'s reasoning over consolidated trajectories}
\begin{wrapfigure}[17]{r}{0.4\linewidth}
  \vspace{-0.3cm}
  \centering
  \includegraphics[width=\linewidth]{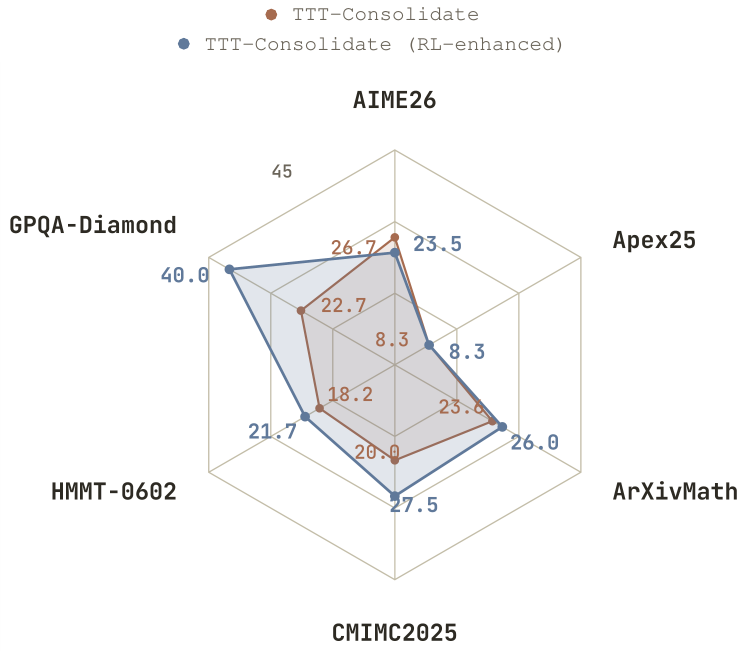}
  \caption{\textbf{\parallelttt{} enhanced with RLVR} on \textit{None} tier trajectories, shows consistent gains over \parallelttt{} trained with SFT.}
  \label{fig:grpo_radar}
\end{wrapfigure}
Reinforcement learning with verifiable rewards (RLVR) has become a standard recipe for eliciting stronger reasoning from LLMs~\citep{guo2025deepseek}. A natural question is whether such reward-driven optimization can also improve \emph{reasoning consolidation}.
We cast RLVR as the outer-loop task for meta-learning over \parallelttt{} (Qwen3-1.7B) where the inner loop performs consolidation, and the outer loop uses GRPO to maximize the likelihood that the consolidated policy reaches a correct answer~\citep{shao2024deepseekmath}.

Figure~\ref{fig:grpo_radar} compares \parallelttt{}~(RL-enhanced) against the supervised \parallelttt{} on aggregating \textit{None} tier trajectories. RL training lifts the average accuracy by 4.6 absolute gain (a $\sim22\%$ relative improvement), with the largest gains on GPQA-Diamond, nearly doubling from 22.7 to 40.0 ($+17.3$). RL improves the final adapted policy's ability to reason over consolidated trajectories, while its current training gains arise mainly from a stronger initialization for consolidation.

\section{Related Work} \label{sec:related}

\textbf{Reasoning Aggregation.\;}
A common way to combine multiple LLM solutions is rule-based voting: self-consistent decoding \citep{wang2022self, brown2024large} samples multiple reasoning trajectories and returns the most frequent answer, with later variants adapting the procedure through dynamic sampling or heuristic filtering \citep{aggarwal-etal-2023-lets, wang2025think, chen2025lspo}. This strategy breaks down when correct solutions occupy minority modes, causing voting to amplify errors rather than correct them. Learned models offer an alternative: reward models score and rank candidates \citep{xu2025acesearcher, yu-etal-2025-self}, Universal Self-Consistency \citep{chen2023universal} has the model inspect all sampled answers and select the most consistent one, and Zhao et al.~\citep{zhao2025majority} instead train the model to reason over and synthesize across candidates as an explicit skill. However, these methods all concatenate candidates directly into the context, suffering from attention dilution and rising cost as context grows. In contrast, \parallelttt{} meta-learns to proactively internalize cross-trajectory information rather than read over the candidate solutions.

\textbf{Parallel Thinking.\;}
Parallel thinking has emerged as a promising paradigm for overcoming the fragility of single-trajectory sequential reasoning \citep{yao2023tree,wang2025survey,liu2025thoughtbubbles,guo2026ope,wu2025thinking}. Most recent work trains models to natively generate and manage multiple thought branches concurrently: Parathinker~\citep{wen2025parathinker} generates diverse, independent thoughts end-to-end and aggregates them into a final answer, while Parallel-R1 \citep{zheng2025parallel} first instills this ability via SFT on easy math problems (cold-start) before extending it to harder problems through reinforcement learning. Other approaches instead refine the structure of parallel reasoning branches \citep{pan2025learning,yang2025multiverse}; notably, Venkatraman et al.~\citep{venkatraman2025recursive} frame the process as an evolutionary loop that recursively aggregates and refines populations of candidate solutions across iterations. Our method differs by decoupling generating parallel thoughts from reasoning over them, enabling the aggregation of thoughts from different sources beyond self-generated ones.

\textbf{Test-Time Training.\;}
MAML~\citep{finn2017model} introduces optimization-based, model-agnostic meta-learning for efficient downstream adaptation; \citet{chen2023reckoning} apply MAML to GPT models, internalizing context knowledge via inner-loop gradient updates before answering queries in the outer loop. A related line of work instead adds or repurposes weight modules that mimic global sequence modeling while avoiding sequential dependency and quadratic scaling~\citep{dalal2025one,feng2026place,zhao2023pitfalls}, such as Sun et al.'s~\citep{sun2024learning} RNN layers with MLP-based hidden states, and new architectures combining persistent neural memory with standard attention to extend context length while preserving context-grounding~\citep{behrouz2025atlas,behrouz2024titans}. Our method, \parallelttt{}, defines reasoning aggregation as a test-time learning process.

\section{Conclusion}

We present \parallelttt{}, a meta-learned reasoning aggregation that actively consolidates thinking trajectories using test-time learning. Rather than reading trajectories as context, \parallelttt{} consolidates them in a lightweight LoRA adapter through a learned adaptation process, enabling the adapted LoRA to synthesize a more reliable solution.
To enable training and evaluation at scale, we introduce \textit{OpenParallelThinking}, a corpus of 13K problems paired with 63K trajectory groups spanning diverse models and three quality levels.
Across ten challenging math and STEM benchmarks, \parallelttt{} outperforms the best aggregation baselines by 20\% on average, with particularly notable gains when all input trajectories are incorrect, confirming that \parallelttt{} can recover from unsuccessful trajectories and arrive at an improved solution through aggregation as test-time learning.
Further analysis shows that \parallelttt{} resists majority-voting shortcuts, scales positively with weak trajectory count, and that RL further improves the adapted policy's ability to reason over consolidated trajectories.

\subsubsection*{Acknowledgments}
We gratefully acknowledge the support of the Swiss National Science Foundation (No. 215390), the AI2050 program at Schmidt Sciences (Grant \#G-25-69783), Sony Group Corporation, and the Swiss National Supercomputing Center (CSCS) in the form of an infrastructure engineering and development project. This research is with support from Google.org and the Google Cloud Research Credits program for the Gemini Academic Program. This project is also funded by the European Union (ERC, RESPECT-LM, 101222478). Views and opinions expressed are however those of the author(s) only and do not necessarily reflect those of the European Union or the European Research Council. Neither the European Union nor the granting authority can be held responsible for them.

\bibliographystyle{plainnat}
{\ \bibliography{custom}}

\clearpage
\appendix
\section{Compute, Latency, and Efficiency Analysis}
\label{app:compute}

We verify that \parallelttt{}'s gains come from its consolidation mechanism rather than simply more test-time computation. We first give a theoretical complexity analysis of \parallelttt{} against majority voting and in-context aggregation, then present empirical evidence against compute-matched baselines.

\subsection{Theoretical Complexity Analysis}
\label{app:compute:theory}

Let $K$ denote the number of intermediate trajectory samples, $L$ the approximate number of tokens in each trajectory, and $T$ the length of the final generated solution. We analyze scaling with $K$ while holding the model, $L$, and $T$ fixed. Under a token-proportional computation model, sampling $K$ trajectories requires $\Theta(KL)$ total computation (\Cref{tab:rb_complexity}). Majority voting therefore has a wall-clock advantage when parallel hardware is sufficient: all trajectories can be generated in parallel, while voting cost is negligible. However, this does not make additional trajectories free. Generating $K$ trajectories concurrently still requires $\Theta(KL)$ total model computation. \parallelttt{} has the same linear asymptotic growth in total computation but a less favorable latency profile, because each consolidation update is sequentially dependent on the updated model state. Its advantage is accuracy per unit of total computation; we do not claim that \parallelttt{} dominates majority voting on every system dimension, but it is substantially more effective when total computation or serving cost is important, because equal or greater computation spent on additional rollouts does not recover its performance.

\begin{table}[h]
  \centering
  \setlength{\tabcolsep}{4pt}
  \renewcommand{\arraystretch}{1.22}
  \caption{\textbf{Asymptotic scaling of aggregation methods} with the number of sampled trajectories $K$.  \parallelttt{} matches majority voting's linear compute in $K$, avoids the quadratic cost of in-context aggregation, and pays only in sequential latency. \vspace{0.2cm}}
  \label{tab:rb_complexity}
  \resizebox{0.95\textwidth}{!}{%
  \begin{tabular}{@{} l p{5.8cm} p{5.8cm} @{}}
    \toprule
    \textbf{Method} & \textbf{Total compute as $K$ increases} & \textbf{Latency, sufficient parallel hardware} \\
    \midrule
    Majority voting & $\Theta(KL)$ sampling $+$ $\Theta(K)$ voting $=$ $\Theta(KL)$ & $\Theta(L)$, independent of $K$ with concurrent generation \\
    \parallelttt{} & $\Theta(KL)$ sampling $+$ $\Theta(KL)$ consolidation $+$ $\Theta(T)$ generation $=$ $\Theta(KL+T)$ & Sampling parallelizes; consolidation is sequential, $\Theta(KL)$, followed by $\Theta(T)$ \\
    In-context aggregation & $\Theta(KL)$ sampling $+$ $\Theta((KL)^2)$ prefill $+$ $\Theta(TKL+T^2)$ decode $=$ $\Theta(K^2)$ w.r.t.\ $K$ & Sampling parallelizes; aggregation cost grows with context length \\
    \bottomrule
  \end{tabular}%
  }
\end{table}

\subsection{Empirical Evidence at Matched Compute}
\label{app:compute:empirical}

Section~\ref{sec:analysis:compute} reports \parallelttt{}'s accuracy against majority voting and in-context aggregation at matched compute (\Cref{tab:main_isocompute,tab:compute_vs_accuracy}). Here we extend that comparison to reward-guided and recursive-refinement baselines, and to inference latency and memory.

We also compare against three additional baselines at matched compute, using \parallelttt{} @8's budget of 1.050e15 total FLOPs: reward-model Best-of-N~\citep{cobbe2021training} using AceMath-7B-RM~\citep{liu2025acemath}, PRM reranking using Qwen2.5-Math-PRM-7B~\citep{lightman2023let}, and Recursive Self-Aggregation~\citep{venkatraman2025recursive}. Recursive refinement receives $N{=}8$ candidate solutions and runs $T{=}2$ rounds; in each round, it draws $N{=}8$ subsets of $k{=}4$ candidates and generates one new candidate per subset, replacing the population. \Cref{tab:rb_isocompute_strong} shows that \parallelttt{} outperforms every baseline at equal compute, both for \textit{None} and \textit{Minority} tiers.

\begin{table}[t]
  \centering
  \scriptsize
  \setlength{\tabcolsep}{4pt}
  \renewcommand{\arraystretch}{1.22}
  \caption{\textbf{Iso-compute comparison against reward-guided and recursive-refinement baselines.} All methods run at \parallelttt{}@8's budget ($\approx$1.05e15 pipeline FLOPs). Best-of-N and PRM reranking select from $N{=}10$ sampled solutions. Recursive Refinement runs $T{=}2$ rounds, each regenerating $N{=}8$ candidates from subsets of $k{=}4$. Reranking scores 0.0 on \textit{None} because it can only select an answer that is already in the pool, and every input there is wrong. \parallelttt{} outperforms the strongest baseline on \textit{Minority} and \textit{None}. \vspace{0.2cm}}
  \label{tab:rb_isocompute_strong}
  \resizebox{0.93\linewidth}{!}{
    \begin{tabular}{@{} l l r r r @{}}
      \toprule
      \textbf{Method} & \textbf{Setting} & \textbf{Pipeline FLOPs} & \textbf{\textit{Minority}} & \textbf{\textit{None}} \\
      \midrule
      Best-of-N, AceMath-7B-RM & $N{=}10$ & 1.040e15 & 24.5 & 0.0 \\
      PRM rerank, Qwen2.5-Math-PRM-7B & $N{=}10$, last step & 1.040e15 & 29.1 & 0.0 \\
      Recursive Refinement & $N{=}8, k{=}4, T{=}2$ & 1.048e15 & 17.9 & 5.4 \\
      \rowcolor{orange!15}
      \textbf{\parallelttt{}} & $K{=}8$ & \textbf{1.050e15} & \textbf{35.4} & \textbf{23.6} \\
      \bottomrule
    \end{tabular}
  }
\end{table}

\emph{Latency and memory.} We measure inference latency using Qwen3-1.7B on a single H100 GPU. All methods receive the same precomputed trajectories, so the latency for sampling trajectories is not counted as part of the total latency since all methods share this measurement. For \parallelttt{}, ``Consolidate'' denotes the complete test-time learning stage that consolidates all trajectories through LoRA updates. \Cref{tab:rb_latency} shows that \parallelttt{} achieves the lowest latency compared to the aggregation baselines, being $2.18\times$ faster than the strongest baseline ($25.238$s vs.\ $55.038$s for Prompt Aggregation in Thinking mode) while improving \textit{Minority} accuracy by 9.6 points and \textit{None} accuracy by 10.6 points. \parallelttt{} has higher peak GPU memory because test-time learning requires gradients and optimizer states. However, it uses a substantially smaller KV cache at decode time because the consolidated model generates from the 156-token problem prompt rather than attending to all concatenated trajectories; this also avoids the growth in KV memory incurred by in-context aggregation as the number of trajectories increases.
The gain comes from \parallelttt{}'s consolidation mechanism, not more test-time computation. Giving the baselines more compute still does not reproduce \parallelttt{}'s performance.

\begin{table}[t]
  \centering
  \setlength{\tabcolsep}{4pt}
  \renewcommand{\arraystretch}{1.22}
  \caption{\textbf{Inference runtime and memory usage on a single H100 GPU.} Trajectory sampling is shared by all methods and excluded from runtime. Baselines decode from a ${\sim}$6.2K-token prompt of concatenated trajectories. \parallelttt{} instead consolidates the trajectories into LoRA weights (1.9\,s) and decodes from the 156-token problem alone. \parallelttt{} has the lowest total latency and the smallest KV cache, while achieving the highest accuracy on both tiers. The cost is ${\sim}$6\,GB higher peak memory for gradients and optimizer states.}
  \label{tab:rb_latency}
  \resizebox{\textwidth}{!}{%
  \begin{tabular}{@{} l l r r r r r r r r r r @{}}
    \toprule
     & & & & \multicolumn{4}{c}{\textbf{Runtime}} & \multicolumn{2}{c}{\textbf{Memory}} & \multicolumn{2}{c}{\textbf{Accuracy}} \\
    \cmidrule(lr){5-8}\cmidrule(lr){9-10}\cmidrule(lr){11-12}
    \textbf{Method} & \textbf{Mode} & \textbf{Out.\ budget} & \textbf{Prompt} & \textbf{Consolidate} & \textbf{Decode} & \textbf{Total} & \textbf{Speed} & \textbf{Peak KV} & \textbf{Peak mem} & \textbf{\textit{Minority}} & \textbf{\textit{None}} \\
    {\footnotesize\itshape} & {\footnotesize\itshape} & {\small\itshape tok} & {\footnotesize\itshape tok} & {\footnotesize\itshape s $\downarrow$} & {\footnotesize\itshape s $\downarrow$} & {\footnotesize\itshape s $\downarrow$} & {\footnotesize\itshape ms/tok $\downarrow$} & {\footnotesize\itshape GB $\downarrow$} & {\footnotesize\itshape GB $\downarrow$} & {\footnotesize\itshape \% $\uparrow$} & {\footnotesize\itshape \% $\uparrow$} \\
    \midrule
    Prompt agg & CoT      & 8{,}192  & 6{,}182 & --    & 25.867 & 25.867 & 3.158 & 1.649 & 15.742 & 17.2 & 4.0 \\
    SFT agg    & CoT      & 8{,}192  & 6{,}182 & --    & 25.867 & 25.867 & 3.158 & 1.649 & 15.742 & 10.2 & 3.2 \\
    Prompt agg & Thinking & 16{,}384 & 6{,}178 & --    & 55.038 & 55.038 & 3.359 & 2.588 & 15.761 & 25.8 & 13.0 \\
    RL agg     & Thinking & 16{,}384 & 6{,}178 & --    & 55.034 & 55.034 & 3.359 & 2.588 & 15.742 & 22.0 & 3.8 \\
    \rowcolor{orange!15}
    \textbf{\parallelttt{}} & CoT & 8{,}192 & \textbf{156} & 1.946 & 23.291 & \textbf{25.238} & \textbf{2.843} & \textbf{0.957} & 21.962 & \textbf{35.4} & \textbf{23.6} \\
    \bottomrule
  \end{tabular}%
  }
\end{table}

\section{Ablation Studies}
\label{app:section:ablation}

\begin{wraptable}[15]{r}{0.5\textwidth}
  \centering
  \vspace{-0.4cm}
  \setlength{\tabcolsep}{3pt}
  \renewcommand{\arraystretch}{1.22}
  \caption{\textbf{Comparing different consolidation methods trained with frontier supervision.} SFT Aggregation (CoT) and \parallelttt{}-SFT train on the same set of identical questions, trajectory groups, and frontier-generated reference solutions. \parallelttt{}-SFT's positive performance gain improves with model scale. Its improvement therefore comes from the meta-learned test-time training, not simply frontier supervision.}
  \label{tab:rb_sft_gain}
  \resizebox{\linewidth}{!}{
    \begin{tabular}{@{} l l r r r @{}}
      \toprule
      \textbf{Model} & \textbf{Trajc} & \textbf{SFT Agg} & \textbf{\parallelttt{}} & \textbf{Gain} \\
      \midrule
      Qwen3-1.7B & None     & 3.2  & \textbf{23.6} & \textbf{+20.4} \\
      Qwen3-1.7B & Minority & 10.2 & \textbf{35.4} & \textbf{+25.2} \\
      Qwen3-4B   & None     & 7.7  & \textbf{40.8} & \textbf{+33.1} \\
      Qwen3-4B   & Minority & 15.4 & \textbf{57.6} & \textbf{+42.2} \\
      \bottomrule
    \end{tabular}
  }
\end{wraptable}
\subsection{Joint Bi-Level Optimization is Necessary}
\label{app:ablation:bilevel}

We verify that \parallelttt{}'s improvement arises from its bi-level consolidation mechanism itself, rather than simply from the frontier-model supervision used in its outer loop. SFT Aggregation in CoT mode is trained with the same questions, trajectory groups, and frontier-generated target solutions as \parallelttt{} (SFT), so both methods receive identical supervised examples and expert supervision. \Cref{tab:rb_sft_gain} compares the two methods on the 10 benchmarks from \Cref{tab:main}: \parallelttt{} consistently and substantially outperforms direct supervised aggregation.
\parallelttt{} uses bi-level optimization with two nested loops, where the inner loop is optimized by the outer loop. We isolate each loop's contribution in turn.

\textbf{Without Inner Loop.} If the inner loop is removed, training reduces to either (1) SFT Aggregation, already shown above, or (2) SFT on math problems with a single CoT trace from the frontier teacher models (same frontier supervision as \parallelttt{}), where sampled trajectories are ignored. \Cref{tab:rb_outer_only} reports this scenario for the \textit{None} tier, both with and without aggregation at inference time. \parallelttt{} outperforms both inference settings under this SFT-trained baseline, with strong performance gains.

\textbf{Without Outer Loop.} If the outer loop is removed, meta-learning is removed, and the inner-loop-only ablation applies the inner-loop update at test time from the default, untrained LoRA initialization --- this is plain test-time training (TTT) without meta-learning. \Cref{tab:rb_inner_only} shows that this ablation performs much worse than the complete \parallelttt{} under the same evaluation setting.

Together, these two ablations show that direct outer-loop supervision (i.e., SFT) is insufficient, while inner-loop updates without meta-learning are also ineffective. \parallelttt{}'s gains arise from optimizing the two components jointly in a bi-level optimization framework, so that the outer loop learns how the inner-loop adaptation should consolidate trajectories.

\begin{table}[t]
  \centering
  \setlength{\tabcolsep}{4pt}
  \renewcommand{\arraystretch}{1.15}
  \caption{\textbf{Outer-loop-only ablation.} We remove the inner loop, reducing training to SFT on a single frontier-teacher CoT trace, ignoring sampled trajectories, evaluated with and without aggregation at inference. The outer loop alone is insufficient. SFT on frontier supervision does not match \parallelttt{}'s performance. \vspace{0.2cm}}
  \label{tab:rb_outer_only}
  \resizebox{0.55\textwidth}{!}{
  \begin{tabular}{@{} l r r r r @{}}
    \toprule
     & \textbf{AIME26} & \textbf{Apex25} & \textbf{ArxivMath} & \textbf{HMMT2602} \\
    \midrule
    \multicolumn{5}{@{}l}{\textit{No aggregation at inference}} \\
    \quad SFT-1.7B & 3.3  & 0    & 5.6  & 3.0 \\
    \quad SFT-4B   & 14.2 & 0    & 11.1 & 6.1 \\
    \multicolumn{5}{@{}l}{\textit{With aggregation at inference}} \\
    \quad SFT-1.7B & 0.8  & 0    & 0.7  & 1.5 \\
    \quad SFT-4B   & 3.3  & 0    & 2.4  & 3.0 \\
    \midrule
    \rowcolor{orange!15}
    \multicolumn{5}{@{}l}{\textit{\parallelttt{} (test-time consolidation)}} \\
    \rowcolor{orange!15}
    \quad \parallelttt{}-1.7B & \textbf{26.7} & \textbf{16.7} & \textbf{23.6} & \textbf{18.2} \\
    \rowcolor{orange!15}
    \quad \parallelttt{}-4B   & \textbf{39.1} & \textbf{33.3} & \textbf{37.9} & \textbf{48.5} \\
    \bottomrule
  \end{tabular}
  }
\end{table}

\begin{table}[t]
  \centering
  \setlength{\tabcolsep}{4pt}
  \renewcommand{\arraystretch}{1.22}
  \caption{\textbf{Inner-loop-only ablation.} We start each problem's test-time update from an untrained LoRA initialization, without outer-loop meta-learning. Plain test-time training fails to consolidate trajectories, so both loops under the bilevel setup are needed. \vspace{0.2cm}}
  \label{tab:rb_inner_only}
  \resizebox{0.7\textwidth}{!}{%
  \begin{tabular}{@{} l l r r r r @{}}
    \toprule
    \textbf{Backbone} & \textbf{Setting} & \textbf{AIME26} & \textbf{Apex25} & \textbf{ArxivMath} & \textbf{HMMT2602} \\
    \midrule
    \multirow{2}{*}{Qwen3-1.7B (None)} & Only inner     & 5.0  & 0 & 10.8 & 3.8 \\
                                 & \cellcolor{orange!15}Inner + outer  & \cellcolor{orange!15}\textbf{26.7} & \cellcolor{orange!15}\textbf{16.7} & \cellcolor{orange!15}\textbf{23.6} & \cellcolor{orange!15}\textbf{18.2} \\
    \multirow{2}{*}{Qwen3-1.7B (Minority)} & Only inner     & 6.7  & 4.2 & 9.0  & 4.6 \\
                                 & \cellcolor{orange!15}Inner + outer  & \cellcolor{orange!15}\textbf{30.0} & \cellcolor{orange!15}\textbf{25.0} & \cellcolor{orange!15}\textbf{34.7} & \cellcolor{orange!15}\textbf{27.3} \\
    \multirow{2}{*}{Qwen3-4B (None)}   & Only inner     & 14.2 & 0 & 12.2 & 13.6 \\
                                 & \cellcolor{orange!15}Inner + outer  & \cellcolor{orange!15}\textbf{39.1} & \cellcolor{orange!15}\textbf{33.3} & \cellcolor{orange!15}\textbf{37.9} & \cellcolor{orange!15}\textbf{48.5} \\
    \multirow{2}{*}{Qwen3-4B (Minority)}   & Only inner     & 15.0 & 0 & 14.9 & 9.1 \\
                                 & \cellcolor{orange!15}Inner + outer  & \cellcolor{orange!15}\textbf{47.6} & \cellcolor{orange!15}\textbf{41.6} & \cellcolor{orange!15}\textbf{42.9} & \cellcolor{orange!15}\textbf{57.7} \\
    \bottomrule
  \end{tabular}%
  }
\end{table}

\subsection{Isolating the Source of Improvement}
\label{app:ablation:controls}

We rule out three alternative explanations for \parallelttt{}'s improvement that do not require genuine cross-trajectory consolidation: repeated exposure to the problem, generic task-distribution activation, and gains available from the learned pre-adaptation initialization alone. \Cref{tab:rb_five_conditions} reports matched controls for each. We additionally isolate cross-trajectory interaction directly with a \textbf{Single-Trace$\times$8} baseline, where we independently adapt on a single trajectory in the group, obtaining 8 different LoRAs, each generating one final solution, and apply majority voting to determine the final answer. This condition uses all eight trajectories and similar total adaptation compute as full \parallelttt{}, and produces eight final solutions. However, information cannot be shared across trajectories before generating the solutions.

\begin{figure}[t]
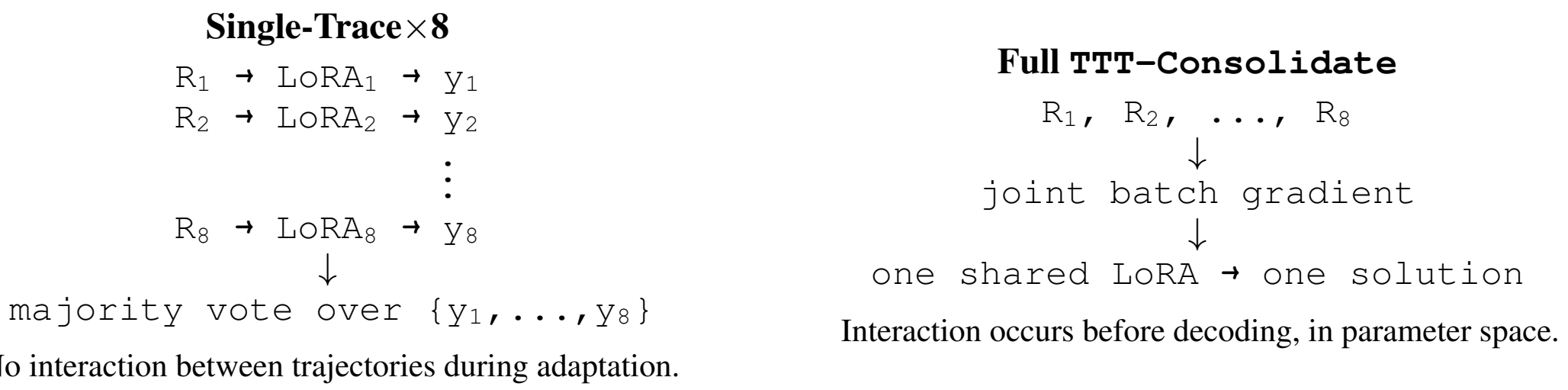

  \centering
  \small
  \begin{minipage}{0.46\linewidth}
    \centering
    \textbf{Single-Trace$\times$8} \\[2pt]
    \texttt{R\textsubscript{1} → LoRA\textsubscript{1} → y\textsubscript{1}}\\
    \texttt{R\textsubscript{2} → LoRA\textsubscript{2} → y\textsubscript{2}}\\
    \texttt{\ \ \ \ \ \ \ \ \ \ \ \vdots}\\
    \texttt{R\textsubscript{8} → LoRA\textsubscript{8} → y\textsubscript{8}}\\
    $\downarrow$ \\
    \texttt{majority vote over \{y\textsubscript{1},\ldots,y\textsubscript{8}\}} \\[4pt]
    {\footnotesize No interaction between trajectories during adaptation.}
  \end{minipage}%
  \hfill
  \begin{minipage}{0.46\linewidth}
    \centering
    \textbf{Full \parallelttt{}} \\[2pt]
    \texttt{R\textsubscript{1}, R\textsubscript{2}, \ldots, R\textsubscript{8}}\\
    $\downarrow$ \\
    \texttt{joint batch gradient} \\
    $\downarrow$ \\
    \texttt{one shared LoRA → one solution} \\[4pt]
    {\footnotesize Interaction occurs before decoding, in parameter space.}
  \end{minipage}
  \caption{
    \textbf{Isolating where trajectory information can interact: Single-Trace$\times$8 vs.\ full \parallelttt{}.} Single-Trace$\times$8 adapts one LoRA per trajectory and combines independent answers from the 8 LoRAs only by majority voting. \parallelttt{} aggregates all trajectories into one shared LoRA before decoding a single final solution.
  }
  \label{fig:rb_interaction_diagram}
\end{figure}

\begin{table}[h]
  \centering
  \setlength{\tabcolsep}{4pt}
  \renewcommand{\arraystretch}{1.22}
  \caption{\textbf{Isolating the source of \parallelttt{}'s improvement.} Every ablation control falls far below full \parallelttt{}. Our method's performance gain requires jointly consolidating the target problem's trajectories during adaptation. \vspace{0.2cm}}
  \label{tab:rb_five_conditions}
  \resizebox{\textwidth}{!}{%
  \begin{tabular}{@{} l p{4.2cm} r r r p{4.2cm} @{}}
    \toprule
    \textbf{Condition} & \textbf{Inner-loop input} & \textbf{Avg.\ acc.} & $\Delta$\textbf{ pre-adapt.} & $\Delta$\textbf{ post-adapt.} & \textbf{Isolate} \\
    \midrule
    Pre-adaptation & No inner-loop update & 7.5\% & --     & $-14.5$ & Initialization alone \\
    Passive repetition & 8 copies of the target question & 5.4\% & $-2.1$ & $-16.6$ & Repeated exposure \\
    Task activation & Trajectory groups from other problems, same task distribution & 4.0\% & $-3.5$ & $-18.0$ & Task-distribution activation w/o target-specific info \\
    Single-Trace$\times$8 & One trajectory per LoRA, 8 LoRAs, majority vote & 11.7\% & $+4.2$ & $-10.3$ & Absence of cross-trajectory interaction \\
    \rowcolor{orange!15}
    \textbf{\parallelttt{}} & All 8 target trajectories, jointly & \textbf{22.0\%} & \textbf{$+14.5$} & \textbf{--} & \textbf{--} \\
    \bottomrule
  \end{tabular}%
  }
\end{table}

Single-Trace$\times$8 performs much worse than \parallelttt{}, showing that independently encoding each trajectory and voting over the generated answers cannot reproduce \parallelttt{}'s improvement. The other controls also perform far below \parallelttt{}, showing that passive question repetition and task-distribution activation do not contribute to \parallelttt{}'s improvement. (To make the 8$\times$ Single-Trace baseline a fair comparison to \parallelttt{}, we additionally sample 8 CoTs from the adapted model after the inner loop and perform a majority vote of these, which yields a $\sim$1\% average additional improvement.)

\parallelttt{} requires information from multiple target-specific trajectories to be jointly consolidated in the same adapted LoRA before it generates the final solution. We accordingly scope the weak-to-strong generalization claim to: \textbf{\parallelttt{} can recover from unsuccessful trajectories and arrive at an improved solution through aggregation as test-time learning.}

\subsection{Role of Reinforcement Learning in Consolidation}
\label{app:ablation:rl_role}

Because RL jointly optimizes the LoRA initialization \emph{and} the model's reasoning after inner-loop updates, the effects of initialization and consolidation are coupled in the post-adaptation measurement. To directly separate performance available in the initialization from performance obtained after trajectory consolidation, we evaluate the same RL-enhanced \parallelttt{} model with and without consolidation at inference time (\Cref{tab:rb_rl_checkpoint}).

\begin{table}[t]
  \centering
  \small
  \setlength{\tabcolsep}{4pt}
  \renewcommand{\arraystretch}{1.22}
  \caption{\textbf{\parallelttt{} model with and without test-time adaptation} on four \textit{None}
  tier benchmarks. Inner-loop adaptation improves both the SFT and RL models on every benchmark. Performance gains
  do not come from the initialization alone. \vspace{0.2cm}}
  \label{tab:rb_rl_checkpoint}
  \resizebox{0.9\textwidth}{!}{
  \begin{tabular}{@{} l l r r r r r @{}}
    \toprule
     & & \textbf{AIME26} & \textbf{Apex25} & \textbf{ArxivMath} & \textbf{HMMT2602} & \textbf{GPQA} \\
    \midrule
    \multirow{3}{*}{\parallelttt{}-SFT}
      & Before TTT & 16.7 & 0 & 2.8  & 6.1  & 6.6 \\
      & After TTT  & 26.7 & 16.7 & 23.6 & 18.2 & 22.7 \\
      & Gain from adaptation    & +10.0 & +16.7 & +20.8 & +12.1 & +16.1 \\
    \midrule
    \multirow{3}{*}{\parallelttt{}-RL}
      & Before TTT & 16.7 & 0 & 4.2  & 9.1  & 19.7 \\
      & After TTT  & 23.5 & 16.7 & 26.0 & 21.7 & 40.0 \\
      & Gain from TTT    & +6.8 & +16.7 & +21.8 & +12.6 & +20.3 \\
    \bottomrule
  \end{tabular}}
\end{table}

The only difference between the two rows within each checkpoint is whether the inference includes the inner-loop adaptation. The inner-loop adaptation clearly improves every benchmark, so the final \parallelttt{} model does not obtain its performance from the initialization alone; its test-time updates contribute a substantial improvement. \parallelttt{}-RL is initialized from the \parallelttt{}-SFT policy before GRPO training, following prior art, so it does inherit frontier supervision through this initialization. To isolate RL's own contribution, we additionally evaluate a compute- and initialization-matched RL Aggregation baseline that initializes from SFT Aggregation and applies the same GRPO training, verifiable reward, rollout budget, and RL update budget used for \parallelttt{}-RL (\Cref{tab:rb_matched_rl}).

\begin{table}[t]
  \centering
  \setlength{\tabcolsep}{4pt}
  \renewcommand{\arraystretch}{1.22}
  \caption{\textbf{Matched SFT$\to$GRPO training path for RL Aggregation and \parallelttt{}}. Both methods get the same frontier-supervised initialization and GRPO budget. RL Aggregation stays behind; thus, RL alone does not explain \parallelttt{}'s improvement.\vspace{0.2cm}}
  \label{tab:rb_matched_rl}
  \resizebox{0.95\textwidth}{!}{%
  \begin{tabular}{@{} l r r r r r @{}}
    \toprule
    \textbf{Method} & \textbf{AIME26} & \textbf{ArXivMath} & \textbf{Apex25} & \textbf{HMMT-2602} & \textbf{Avg} \\
    \midrule
    RL Aggregation: Base $\to$ GRPO (base-initialized) & 6.7  & 2.8  & 0.0 & 0.4 & 2.48 \\
    RL Aggregation: SFT $\to$ GRPO (matched)    & 10.0 & 1.2  & 0.0 & 9.1 & 5.08 \\
    \rowcolor{orange!15}
    \parallelttt{}-SFT $\to$ GRPO (\parallelttt{}-RL) & \textbf{23.5} & \textbf{26.0} & \textbf{16.7} & \textbf{21.7} & \textbf{21.98} \\
    \bottomrule
  \end{tabular}%
  }
\end{table}

This comparison controls for both frontier-supervised initialization and RL training; \parallelttt{} still outperforms this baseline under the matched SFT-to-RL learning path. We do not interpret this as showing that RL primarily improves the consolidation process (though this is benchmark-dependent); however, the adaptation gain under the same checkpoint shows that the inner loop remains important. RL primarily strengthens the initialization, but the resulting model still relies substantially on trajectory-conditioned test-time adaptation: \emph{RL improves the final adapted policy's ability to reason over consolidated trajectories, while its current training gains arise mainly from a stronger initialization for consolidation.}

\section{Consolidation Dataset: Construction, Quality Control, and Verification}
\label{app:section:dataset}

\subsection{Evaluation Benchmark and Trajectory Preparation}
\label{app:subsection:reasoning_trajectory}

\begin{wraptable}[20]{r}{0.55\linewidth}
  \vspace{-0.4cm}
  \centering
  \setlength{\tabcolsep}{6pt}
  \renewcommand{\arraystretch}{1.15}
  \caption{\textbf{Summary of the ten reasoning benchmarks.} Each question is used to generate 40 candidate trajectories (5 source models $\times$ 8 samples), from which three quality-stratified trajectory groups are formed.}
  \label{tab:benchmark_stats}
  \small
  \resizebox{\linewidth}{!}{
  \begin{tabular}{@{} l l r @{}}
    \toprule
    \textbf{Benchmark} & \textbf{Domain} & \textbf{\# Questions} \\
    \midrule
    \multicolumn{3}{l}{\textit{Math Reasoning}} \\[1pt]
    \quad AIME 2025           & Competition Math      & 30  \\
    \quad AIME 2026           & Competition Math      & 30  \\
    \quad Apex 2025           & Competition Math      & 12  \\
    \quad ArXiv Math          & Research Math         & 72  \\
    \quad CMIMC 2025          & Competition Math      & 40  \\
    \quad HMMT Feb 2026       & Competition Math      & 33  \\
    \quad HMMT Nov 2025       & Competition Math      & 30  \\
    \quad IMO-Bench           & Olympiad Math         & 400 \\
    \multicolumn{3}{l}{\textit{STEM Reasoning}} \\[1pt]
    \quad GPQA Diamond        & Scientific Reasoning  & 198 \\
    \quad TheoremQA           & STEM Theorem Reasoning & 800 \\
    \midrule
    \textbf{Total}            &                       & \textbf{1,645} \\
    \bottomrule
  \end{tabular}
  }
\end{wraptable}

\paragraph{Benchmark Coverage and Question Pool}
We assemble a pool of 1,645 unique questions drawn from ten challenging reasoning benchmarks spanning competition mathematics and scientific reasoning.
We detail the full pipeline to construct the reasoning consolidation dataset used in our experiments. Table~\ref{tab:benchmark_stats} provides a complete breakdown of each benchmark, its domain, and the number of questions it contributes.
Together, these benchmarks cover a broad difficulty spectrum: from competition-level problems designed for elite high-school contestants (AIME, HMMT, CMIMC) and olympiad-level challenges (IMO-Bench) to research-level mathematics (ArXivMath), PhD-level scientific reasoning across physics, chemistry, and biology (GPQA Diamond), and multi-domain STEM theorem application (TheoremQA).
Notably, the difficulty of these benchmarks substantially exceeds the reach of smaller models, posing a genuine challenge even for capable reasoning models.

\paragraph{Source Models and Trajectory Generation}
For each question in the pool, we sample 8 independent solution attempts from each of five source models, yielding 40 candidate trajectories per question.
The five source models span a range of scales and capabilities.
\textbf{Qwen2.5-1.5B-Instruct}, \textbf{Qwen3-1.7B}, and \textbf{Qwen3-4B} represent lightweight to mid-range open models.
\textbf{Qwen3-30B-A3B} is a sparse mixture-of-experts model with 30B total but only 3B active parameters per token, offering strong reasoning at moderate inference cost.
\textbf{Gemini 2.0 Flash} is a capable closed-source model serving as the strongest single contributor.
The first four open-source models operate in extended thinking mode and produce very long internal reasoning traces before writing a final answer. Gemini 2.0 Flash is a closed-source model that outputs only a brief CoT solution without an explicit reasoning trace.

\begin{figure}[b]
  \centering
  \includegraphics[width=0.9\linewidth]{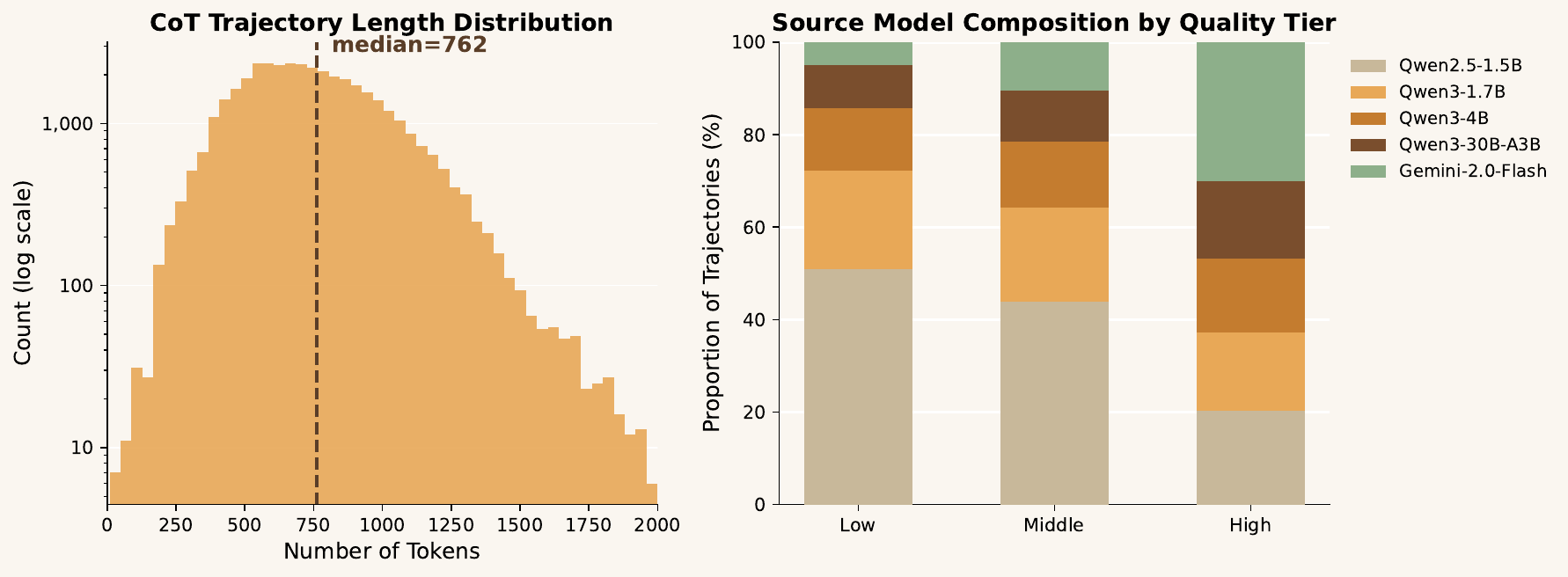}
  \caption{%
    \textbf{Left}: Distribution of extracted CoT lengths over reasoning trajectory collection.
    \textbf{Right}: Source model composition by quality tier, where weaker models dominate the \textit{None} tier while stronger models contribute more in the \textit{Majority} tier.}
  \label{fig:traj_stats}
\end{figure}

\paragraph{CoT Extraction from Thinking Outputs}
For the four open-source thinking models, each raw generation contains a lengthy internal reasoning trace in \texttt{<think>...</think>} tags, which can exceed tens of thousands of tokens and is therefore too long to include directly in aggregation prompts.
We extract only the brief final solution appearing after the thinking block as the CoT trajectory. For Gemini 2.0 Flash, the full output is used directly.
As shown in Figure~\ref{fig:traj_stats} (left), the extracted CoTs have a median length of 762 tokens (25th--75th percentile: 559--927), confirming they are compact enough for multi-trajectory aggregation contexts.

\paragraph{Quality-Based Stratification}
For each question, we sample 8 trajectories from the 40 candidates and assign them to one of three quality tiers based on how many of the 8 are correct:
\begin{itemize}
  \item \textit{None} ($0/8$ correct): all selected trajectories are incorrect, representing the hardest aggregation scenario where the model must extract useful reasoning steps from entirely wrong solutions.
  \item \textit{Minority} ($0$--$3/8$ correct): a mix of correct and incorrect trajectories, reflecting the most common real-world setting where the aggregator must identify and leverage correct solutions among a noisy mixture.
  \item \textit{Majority} ($4$--$7/8$ correct): the majority of selected trajectories are correct, an extreme case testing whether the model can reliably synthesize a consistent final answer.
\end{itemize}

Two boundary cases arise in practice: some questions are so easy that all 40 candidates are correct (pass@40 $= 100\%$), and others are so hard that all 40 are wrong (pass@40 $= 0\%$). For these questions, we use frontier models (GPT-5.3 and Claude~4.6) to manually correct the imbalance with access to the ground-truth answer. In the all-correct case, a subset of trajectories is rewritten to introduce plausible reasoning errors that lead to a wrong final answer. In the all-wrong case, a subset of trajectories is corrected to produce a valid solution. This ensures that every question participates in all three quality tiers, yielding $1{,}645 \times 3 = 4{,}935$ trajectory groups in total.

\paragraph{Source Model Composition across Quality Tiers}
As shown in Figure~\ref{fig:traj_stats} (right), source model composition shifts systematically across quality tiers.
In the \textit{None} tier, 51\% of trajectories come from Qwen2.5-1.5B, the weakest model, because it fails most often.
As quality increases, stronger models contribute more: Gemini 2.0 Flash rises from 5\% (\textit{None}) to 30\% (\textit{Majority}), and Qwen3-30B-A3B from 9\% to 17\%, while Qwen2.5-1.5B's share falls to 20\%.
This emergent shift confirms that model capability and trajectory quality are tightly coupled.

\subsection{\openparallel{} Training Corpus Construction}
\label{app:subsection:openparallel_construction}

This section details the construction pipeline for \openparallel{} summarized in Section~\ref{sec:method}.

\textbf{Source Diversity} \; We pool problems from OmniMath \citep{gao2024omni}, DeepScale \citep{deepscaler2025}, NuminaMath \citep{numina_math_datasets}, and OpenMathInstruct \citep{toshniwal2024openmath2}, yielding 19,615 unique problems after hash-based deduplication.

\textbf{Trajectory Diversity} \; We sample from four solvers spanning different scales, families, and training recipes: R1-7B \citep{deepseekai2025deepseekr1incentivizingreasoningcapability}, Nemotron3-14B \citep{chen2025acereason}, Qwen3-4B, and Qwen3-30B-A3B \citep{yang2025qwen3}, since a homogeneous source would bias the aggregator toward a single reasoning style. We drop the extended reasoning segment of each trajectory with probability 0.5\footnote{For thinking models, the reasoning segment corresponds to the content between two special thinking tokens, most commonly \texttt{<think>} and \texttt{</think>}.}, exposing the aggregator to a realistic mix of detailed and compact reasoning. Trajectories are then randomly grouped per question under a 16K-token budget, so that group size varies naturally with length \citep{zhao2024long} (Figure~\ref{fig:dataset_statistic}, left).

\textbf{Balanced Difficulty} \; We filter trajectory groups by correctness ratio (Figure~\ref{fig:dataset_statistic}, center) to provide more training examples on reasoning aggregation over groups where the majority are incorrect trajectories, which provides stronger learning signals without the majority correct answer shortcut.

\textbf{Final Dataset} \; The final release contains 12,810 questions and 63,507 trajectory groups. We sample 8,000 instances for RL training and reserve the rest for SFT.

\subsection{Verifying Genuine Trajectory Consolidation}
\label{app:subsection:judge}

A natural concern is whether \parallelttt{}'s inner loop learns to memorize input trajectories rather than consolidate useful sub-reasoning, and whether its gains on the \textit{None} tier are an artifact of the frontier-model-rewritten trajectories used to balance that tier (\Cref{app:subsection:reasoning_trajectory}) rather than genuine consolidation of naturally occurring failures. We address both concerns with a construction-based argument and a new LLM-as-judge analysis.

\paragraph{Ruling out memorization by construction.} When all 8 trajectories in the \textit{None} tier are naturally unsuccessful trajectories sampled from source models, each ending in a wrong answer, there are no correct candidates to memorize and recall. Averaged over the ten benchmarks, \parallelttt{} reaches 23.58 vs.\ 13.02 for the strongest baseline (Qwen3-1.7B) and 40.83 vs.\ 19.63 (Qwen3-4B). For \parallelttt{} to output a correct answer in this setting, it must not have only memorized trajectories leading to the correct answer. \Cref{fig:scaling_analysis} also shows that \parallelttt{} generally improves as more incorrect trajectories are added, while baselines stagnate or degrade. Thus, \parallelttt{} does not depend on simply copying the input trajectories' answers.

\paragraph{Ruling out trajectory-editing artifacts.} Edited trajectories are a negligible presence in the \textit{None} setting. Among all 1{,}645 evaluation questions, only two (one from GPQA; one from TheoremQA) contain any rewritten trajectory in the \textit{None} setting; the remaining eight benchmarks contain none. Counting these two questions as incorrect decreases the GPQA and TheoremQA scores by less than 0.5 percentage points and leaves every other benchmark score unchanged. The weak-to-strong gains in \Cref{tab:main} are therefore primarily driven by naturally occurring model failures, not edited rewriting.

\paragraph{Judge Setup} We use three frontier LLM judges, Fable~5, Opus~5, and GPT-5.6-Sol, each run at extra-high reasoning effort. Every sampled solution is independently labeled by all three judges, and the final label is the majority vote across the three.

\paragraph{Synthesis and Artifact-Exploitation Judging (\textit{None})} We sample 100 \parallelttt{} (Qwen3-1.7B) solutions on \textit{None}-tier problems, where every one of the eight input trajectories reaches an incorrect answer. For each solution, the judges are shown the problem, the eight input trajectories, and \parallelttt{}'s final solution, and are asked two binary questions: (1) whether the final solution derives an improved reasoning trajectory from the incorrect inputs (\textit{synthesis}) rather than simply copying one input's reasoning verbatim, and (2) whether the solution exploits stylistic or distributional cues rather than demonstrating mathematical understanding (\textit{artifact exploitation}). 98/100 solutions are labeled as synthesis, and 97/100 show no evidence of artifact exploitation. We additionally audit the two \textit{None}-tier questions containing an edited trajectory (\Cref{app:subsection:reasoning_trajectory}) specifically for exploitable artifacts; the judges find that the edits introduce mathematically meaningful errors without creating clear exploitable cues.

\paragraph{Memorization Judging (\textit{Minority})} We separately sample 100 \parallelttt{} (Qwen3-1.7B) solutions on \textit{Minority}-tier problems, where a subset of the eight input trajectories is already correct. Judges are shown the same problem, trajectory, and solution triple, and asked to label the final solution as \textit{memorization} (near-verbatim reproduction of one input trajectory's reasoning and answer) or \textit{consolidation} (reasoning synthesized or corrected using information across multiple inputs). 1/100 solutions is labeled memorization.

\paragraph{Judge Prompt Template} Table~\ref{tab:judge_prompt_template} shows the prompt template shared across the three judging tasks, with the task-specific question and label set substituted for each protocol above.\footnote{More detailed instructions, including definitions of artifact exploitation, memorization, and consolidation, will be released with the code.}

\begin{table*}[t]
    \caption{Prompt template for the LLM-as-judge analysis on the final solutions produced by \parallelttt{}.}
    \label{tab:judge_prompt_template}
    \centering
    \begin{tabular}{lp{10.2cm}}
    \toprule
    \textbf{Component} & \textbf{Content} \\
    \midrule
    \multirow{6}{*}{\shortstack[l]{Judge Prompt}}
      & You are evaluating whether a model's final solution to a problem reflects \\
      & \{synthesis vs.\ artifact exploitation $\mid$ consolidation vs.\ memorization\}. \\
      & Problem: \{problem\} \\
      & Input trajectories: \{trajectory$_0$\}, \ldots, \{trajectory$_7$\} \\
      & Final solution: \{solution\} \\
      & Question: \{task-specific question\} \\
      & Respond with a single label from \{label set\} and a one-sentence justification. \\
    \bottomrule
    \end{tabular}
\end{table*}

Taken together, these results show clear evidence that \parallelttt{} consolidates multiple naturally sampled unsuccessful trajectories into a stronger, improved solution, rather than memorizing correct answers or exploiting artifacts introduced by trajectory editing.

\section{Training and Inference Details}
\label{app:subsection:implementation_details}
\paragraph{Hardware Information} All experiments are conducted on a single NVIDIA H100 GPU with 93\,GB of HBM on an internal Slurm-managed research cluster. Each job is assigned 64 CPU cores and 64\,GB of system RAM with exclusive GPU access throughout training.

\paragraph{Baseline Training Hyperparameters}
The SFT aggregation model is obtained by full-parameter fine-tuning using LLaMA-Factory \cite{zheng2024llamafactory}. We adopt the no-thinking chat template with a context window of 20,576 tokens. Training runs for 1 epoch with a learning rate of $1\times10^{-5}$, a cosine schedule with 10\% linear warmup, and an effective global batch size of 32. All training is conducted in \texttt{bf16} precision.
For RL aggregation, we implement GRPO~\citep{shao2024deepseekmath} under the Verl \cite{sheng2024hybridflow} framework, with thinking mode enabled. We use a learning rate of $3\times10^{-6}$, a global batch size of 64 prompts, and $n=5$ rollouts per prompt. The maximum prompt and response lengths are set to 8,192 and 16,384 tokens, respectively.

\paragraph{Our Method Training Hyperparameters}
We train \parallelttt{} for up to 1 epoch with early stopping based on validation loss, using PyTorch Lightning\footnote{\url{https://lightning.ai/docs/pytorch/stable/}} modified to support meta-gradients. The outer loop optimizer is AdamW~\citep{loshchilov2017decoupled} with a learning rate of $1\times10^{-5}$, weight decay of 0.01, and a cosine schedule with 3\% warmup steps. The consolidation module is parameterized as a LoRA adapter with rank 256 and $\alpha=128$, applied to all weight matrices of the model. For inner-loop adaptation, we reuse the encoding scheme and hyperparameters from that prior work, including its Truncated Gradient Unrolling (TGU) scheme~\citep{patil2024gorilla}: the first $S-T$ inner steps are executed without graph retention, and only the final $T$ steps are differentiated, reducing memory from $O(S|\phi|)$ to $O(T|\phi|)$.

For the RL-enhanced variant, we introduce GRPO as the outer-loop objective on top of the existing PyTorch Lightning meta-training framework, without adopting dedicated RL infrastructure such as vLLM or VERL. The outer-loop learning rate is reduced to $1\times10^{-6}$ with weight decay 0.003. Rollouts are capped at 2,048 new tokens. To further reduce peak memory from higher-order gradient computation, we set \texttt{unroll\_start}$=3$ with $n_{\text{inner}}=4$, retaining only the final inner step in the gradient graph.

\paragraph{Evaluation Hyperparameters}
All methods use greedy decoding (temperature $= 0$), with the end-of-sequence token taken from the model tokenizer's default. The maximum generation length is 8,192 tokens in CoT mode and 16,384 tokens in thinking mode. Baseline inference is accelerated with vLLM~\citep{kwon2023efficient}, while \parallelttt{} generates answers through the Hugging\,Face Transformers pipeline\footnote{\url{https://huggingface.co/docs/transformers/v4.43.4/en/main_classes/pipelines}}.

\begin{table*}[t]
    \caption{Prompt templates for model-based aggregation baselines and \parallelttt{}.}
    \label{tab:prompt_templates}
    \centering
    \small
    \scalebox{0.95}{
    \begin{tabular}{lp{10.2cm}}
    \toprule
    \textbf{Method} & \textbf{Prompt Template} \\
    \midrule
    \multirow{9}{*}{\shortstack[l]{Prompt Aggregation \\ SFT Aggregation \\ RL Aggregation}}
      & Given the following problem: \{problem\} \\
      & and these solution attempts: \\
      & Solution attempt 0: \{trajectory$_0$\} \\
      & Solution attempt 1: \{trajectory$_1$\} \quad \ldots\ (more attempts) \\
      & Carefully review the provided solutions, using them as starting points, \\
      & correcting mistakes, filling in gaps, and/or combining useful ideas, \\
      & to produce a final, comprehensive, and correct solution to the problem. \\
      & Analyze step by step and put your final answer in the format \textbackslash{}boxed\{\} \\
    \midrule
    \multirow{2}{*}{\shortstack[l]{\parallelttt{} \\ (Inner Loop)}}
      & record$_0$: \{trajectory$_0$\} \\
      & record$_1$: \{trajectory$_1$\} \quad \ldots\ (one sequence per trajectory, $K$ steps total) \\
    \midrule
    \multirow{4}{*}{\shortstack[l]{\parallelttt{} \\ (Outer Loop)}}
      & record$_0$ to record$_{K-1}$ are the information you memorized. \\
      & Question: \{problem\} \\
      & Please solve this problem step by step by reasoning from your memory \\
      & and put your final answer in the format \textbackslash{}boxed\{\}. \textbackslash{}nAnswer: \\
    \bottomrule
    \end{tabular}
    }
\end{table*}

\section{Prompts}
\label{app:subsection:Prompts}

The three model-based aggregation baselines (\textbf{Prompt Aggregation}, \textbf{SFT Aggregation}, and \textbf{RL Aggregation}) share an identical prompt template in training and inference, differing only in the underlying model checkpoint. Prompt Aggregation uses the base model directly, while SFT and RL Aggregation use models further fine-tuned via supervised fine-tuning and reinforcement learning, respectively. The shared prompt concatenates all $K$ solution attempts with indexed headers and instructs the model to synthesize a final correct answer by reviewing them. \parallelttt{} uses two distinct prompts. The inner-loop prompt encodes each reasoning trajectory as a named memory record (\texttt{record\_$i$}) using a causal language modeling objective, compressing the trajectory set into the LoRA adapter weights via gradient updates. The outer-loop prompt then instructs the adapted model to leverage memorized records to answer questions. Table~\ref{tab:prompt_templates} shows the detailed prompt templates.

\section{Limitations}
\label{app:subsection:limitations}

\paragraph{Memory Cost of Meta-RL Training.}
Meta-learning requires backpropagating through the inner optimization loop, which increases memory cost compared to standard fine-tuning. Replacing the SFT outer loss with GRPO further amplifies peak memory usage due to rollout generation and higher-order gradient computation. LoRA substantially reduces this overhead, while we believe further reductions through gradient checkpointing or more aggressive truncation schemes would broaden accessibility to researchers with tighter hardware budgets and enable scaling to larger backbone models.

\paragraph{Scope of Reasoning Domains.}
Our experiments cover mathematical and STEM reasoning, where trajectory quality can be easily and completely assessed through verifiable final answers. Extending \parallelttt{} to agentic settings, where reasoning involves tool use, multi-step environment interaction, and open-ended planning, introduces new challenges in defining trajectory quality and designing appropriate reward signals. These are natural and compelling directions for future work, and the meta-learning formulation of \parallelttt{} is in principle compatible with such settings.

\end{document}